\documentclass[]{rsos}

\usepackage[noend]{algpseudocode}
\usepackage{amsmath}
\usepackage{algorithm}
\usepackage{txfonts}
\usepackage{wrapfig}

\begin{document}

\title{Detecting and Diagnosing Faults in Autonomous Robot Swarms with an Artificial Antibody Population Model}

\author{
James O'Keeffe}

\address{Department of Computer Science, University of York, United Kingdom\\}

\subject{Robotics, Artificial Intelligence, Biomimetics}


\corres{James O'Keeffe\\
\email{james.okeeffe@york.ac.uk}}
\newcommand{\algorithmautorefname}{Algorithm}
\begin{abstract}
An active approach to fault tolerance, the combined processes of fault detection, diagnosis, and recovery, is essential for long term autonomy in robots -- particularly multi-robot systems and swarms. Previous efforts have primarily focussed on spontaneously occurring electro-mechanical failures in the sensors and actuators of a minority sub-population of robots. While the systems that enable this function are valuable, they have not yet considered that many failures arise from gradual wear and tear with continued operation, and that this may be more challenging to detect than sudden step changes in performance. This paper presents the Artificial Antibody Population Dynamics (AAPD) model -- an immune-inspired model for the detection and diagnosis of gradual degradation in robot swarms. The AAPD model is demonstrated to reliably detect and diagnose gradual degradation, as well as spontaneous changes in performance, among swarms of robots of varying sizes while remaining tolerant of normally behaving robots. The AAPD model is distributed, offers supervised and unsupervised configurations, and demonstrates promising scalable properties. Deploying the AAPD model on a swarm of foraging robots undergoing gradual degradation enables the swarm to operate on average at between 70\% - 97\% of its performance in perfect conditions and is able to prevent instances of robots failing in the field during experiments in most of the cases tested.
\end{abstract}

\begin{fmtext}
\section{Introduction}
A significant barrier to the real-world deployment of autonomous robots, particularly in environments that are populated, uncontrolled, difficult to access, and/or safety-critical, is the risk of failure or loss of autonomous control 
\end{fmtext}

\maketitle

in the field. These risks compound for multi-robot systems (MRS), where there is additional vulnerability to faults and failures in the interaction space between agents.

Swarm robotic systems (SRS), a variant of multi-robot systems (MRS), are suited to spatially distributed tasks -- particularly in dangerous/inaccessible environments -- because of their redundancy of hardware and distributed control architectures, meaning that there is no single point of failure \cite{csahin2004swarm}. \c{S}ahin \cite{csahin2004swarm} proposes that these properties provide robot swarms with an innate robustness -- that is, the ability to tolerate faults and failures in individual robots without significant detriment to the swarm as a whole. However, later studies demonstrate that failures in individual robots can significantly disrupt overall swarm performance  -- particularly where a partially failed robot is able to maintain a communication link with other robots and influence their behaviour \cite{winfield2006safety}. Further investigation concludes that an active approach to fault tolerance is necessary if robot swarms are to retain long-term autonomy, and specifically highlights artificial immune systems (AIS) as a promising solution \cite{bjerknes2013fault}.

AIS are derived from observations of the natural immune system \cite{de2002artificial}. One of the defining characteristics of the natural immune system, and one of the most desirable properties for transferring to engineered systems, is its ability to learn and remember infections it has previously encountered and to detect and destroy those infectious cells more efficiently on subsequent encounters. \textit{Maintenance} is defined by Cohen \cite{cohen2000tending} to be the property of the natural immune system that enables it to protect its host against harm it will receive during its life, and comprises three stages: Recognition, cognition, and action. This can be mapped to three stages of active fault tolerance in engineered systems: Fault detection, fault diagnosis, and recovery (FDDR) \cite{christensen2007exogenous, millard2014run, o2018fault}.

Although there is a vast body of literature on FDDR in single robot systems \cite{khalastchi2018fault}, the techniques used are often unsuited for application to SRS because they do not exploit system multiplicity, are not designed with distributed implementation in mind, and are necessarily endogenous -- i.e., a robot must detect faults within itself, which can be problematic if a faulty robot is unable or unwilling to identify its own faulty status \cite{christensen2008fault}. There are a number of recent reviews covering fault detection and diagnosis in MRS and SRS, e.g. \cite{khalastchi2019fault, miller2021survey,bossens2022resilient}, which highlight that previous work towards fault tolerance in SRS/MRS has mostly examined individual elements of FDDR in isolation, with the majority of work being on fault detection. 

Fault detection typically compares observed behaviour with a predefined model of expected behaviour (model-based approaches), or against system models that are built in real-time from observations during operation (data-driven approaches), where discrepancies and outliers indicate potential faults \cite{khalastchi2019fault}. Fault detection in SRS/MRS differs from fault detection in single robots in a few ways -- most notably in the opportunity to exploit the multiplicity of robots to produce a data-driven online model of normal behaviour. A key feature of SRS is their distributed control architectures, meaning that there is no single point of system-wide failure -- an advantageous feature for fault tolerant systems. By contrast, a single robot would typically require a predefined model of normal behaviour in order to detect faults. Producing a comprehensive model of normal behaviour is non-trivial, especially where the environment a robot is deployed in and the behaviours it can exhibit are variable, because of the limited availability of labelled training data. This can be problematic for supervised learning approaches, leading to inflexible systems that are only appropriate for deployment in limited scenarios and conditions. Recent work highlights that unsupervised learning approaches may be necessary for near-term robotic applications \cite{wu2021unsupervised}. Data-driven detection models, by contrast, do not necessarily need to be trained on robot data collected prior to deployment, and have the advantage of being adaptive -- i.e. the SRS can alter its behaviour and, along with it, its implicit model of normal behaviour in real-time. These approaches account for the majority of previous work on fault detection in SRS (e.g. \cite{christensen2009fireflies,lau2011adaptive,khadidos2015exogenous,taraporePLOS,tarapore2019fault,lee2022data,strobel2023robot, carminati2024distributed}). However, they are also limited insofar that they cannot easily detect faults affecting a majority of robots, such as those resulting from external influences or environmental adversity. Faults caused by these factors may be system-wide, but they equally require detection and mitigation strategies if the system is to retain long-term autonomy. A utility for learned models of normal behaviour and model-based approaches to fault detection in SRS, such as the approach taken by Millard et al. \cite{millard2014run, millard2016exogenous}, thus remains. A hybrid model-based and data-driven fault detection mechanism may offer SRS a means of benefitting from the advantages of both -- much like the innate and adaptive components of the natural immune system \cite{owen2013kuby} -- especially if model-based components are able to exploit unsupervised learning.  

Different types of fault cause robots to fail in different ways, degrading performance at the individual robot and swarm levels with varying degrees of severity. Diagnosing a fault allows a system to determine an appropriate resolution and the urgency with which it should be carried out. Where fault detection in SRS can, at its simplest level, provide a binary yes or no response as to whether a fault exists anywhere in the system, fault diagnosis is a means of identifying which robot has the fault, and specifically which of its sensors and actuators is faulty. Previous literature on fault diagnosis in MRS focuses on diagnosing planning or coordination faults or, more pertinent to this work, diagnosing electro-mechanical failures \cite{khalastchi2019fault}. Similar to fault detection, the discrepancies between observed and model-predicted behaviours can also be used to diagnose faults in robots \cite{daigle2007distributed, carrasco2011fault}. The requirement to train diagnostic models can be circumvented with the use of preprogrammed diagnostic tests to isolate the root cause of a fault \cite{kutzer2008toward}. In SRS, O'Keeffe et al. \cite{o2018fault, o2023hardware} offer a hybrid solution that uses preprogrammed diagnostic tests to build a learned online model of faulty behaviour in an unsupervised process that allows system system efficiency to improve over time. This work highlights that the processes of fault detection and diagnosis are intrinsically linked -- to the extent that the consideration of one process is limited without the simultaneous consideration of the other.

One limiting factor of the works towards the detection and diagnosis of electro-mechanical faults in SRS and MRS discussed so far is the types of faults and the fault modelling considered. Where electro-mechanical faults affecting sensor and actuator hardware are examined (e.g. \cite{taraporePLOS,tarapore2019fault, millard2014run, millard2016exogenous, o2017towards, o2017diagnosis, o2023hardware, khadidos2015exogenous, lee2022data, carminati2024distributed}), they are spontaneously `injected' into a minority sub-population of robots while the majority of the SRS remains operational at a uniform level. While modelling faults as spontaneous events is appropriate for some kinds of fault, e.g. a robot that is immobilised suddenly after becoming stuck on an obstacle, other types of fault occur gradually. An example of this would be the accumulation of dust and debris on motor and sensor hardware, highlighted in the real-world study by Carlson et. al. \cite{CarlsonMTBF} as one of the most common causes of failure in the field.

Modelling faults as gradual degradation of robot hardware alters the fault detection problem. Whereas previous work on fault detection in SRS has sought to detect a fault as soon as possible after it has been injected, with a clear demarcation between faulty and non-faulty robots, gradual degradation blurs this line and requires that an optimum region for detecting degradation be found. It is obviously inefficient to allow a robot to degrade excessively before detecting it as faulty, however it is similarly inefficient to detect faults in robots exhibiting only minor reductions in performance. The problem is further complicated by the fact that all robots possess a MTBF and are therefore subject to degradation simultaneously, albeit at varying rates. Each robot degrading independently provides additional challenge to data-driven approaches that rely on a majority of robots providing an implicit model of normal behaviour and removes the guarantee that a majority population will always be `healthy'. Furthermore, a feedback loop between the reliability of data-driven fault detection and the quality of data-driven models of normal behaviour is introduced -- if robots are allowed to degrade for too long, the implicit model of normal behaviour provided by the SRS will be of poorer quality, and it will therefore be harder to reliably detect faults as behavioural deviations from it. These challenges have not yet been addressed in fault tolerant SRS research.

Considering faults occurring through gradual degradation presents an opportunity to implement fault tolerance in SRS as a predictive measure. By detecting early-stage degradation on robot hardware before it reaches a critical failure point, an at-risk robot can be allowed a grace period in which to reach a controlled area for receiving maintenance -- reducing the risk of failure in the field. This is the underlying principle of preventative maintenance, which is widely applied across industrial machinery for its long-term cost saving benefits and reduction of downtime \cite{malik1979reliable}, but has not yet been applied to fault detection in SRS. Part of the reason for this is that such an approach contradicts traditionally held views that SRS are, by their very definition, tolerant to the loss of individuals \cite{csahin2004swarm}. However, it was recently shown \cite{o2025anticipating} that early detection of faults and the prevention of failure in the field provides an advantage to SRS in many scenarios when compared with traditional approaches in which faulty robots are shut down \cite{khadidos2015exogenous} or isolated \cite{strobel2023robot}. 

Having now identified several gaps in fault tolerant SRS literature, this paper presents the Artificial Antibody Population Dynamics Model (AAPD-Model) for the detection and diagnosis of potential faults and hazards. Inspired by Farmer et al.'s \cite{farmer1986immune} model of antibody population dynamics, the AAPD model aims to mimic the self-tolerance and learning properties of the natural immune system. 

The AAPD model is trained and tested on a swarm of simulated TurtleBot3 robots using Robot Operating System (ROS) 2 and Gazebo Classic. SRS of varying sizes (up to 20) are examined, where each robot is subject to stochastic gradual degradation of motor and sensor hardware while the SRS performs some variation of foraging task. The AAPD model is assessed on its ability to detect faults in robots that have degraded below a defined threshold, tolerate robots that operate above the threshold, and to prevent instance of robot failure in the field -- i.e. by detecting faults while robots are still operationally capable of reaching a controlled area. The AAPD model is also tested against existing swarm fault detection and fault diagnosis models in scenarios where a direct comparison can be made.

This paper makes the following key contributions to SRS fault tolerance/fault detection literature:
\begin{itemize}
    \item The AAPD model: An algorithm for the autonomous detection and diagnosis of faults in SRS, inspired by models of the natural immune-system. The AAPD is a hybrid data-driven and model-based approach that is distributed, scalable, and can accommodate supervised and unsupervised configurations. A simplified version of the model is implemented in \cite{o2025anticipating}. However, the AAPD model presented in this article is expanded and achieves novelty through rigorous assessment.
    \item The integration of fault detection and fault diagnosis processes for the first time in SRS. These two intertwined processes have always been examined in isolation in previous work.
    \item The modelling of robot faults as gradual stochastic degradation in sensor and actuator hardware. Despite gradual degradation being one of the most common causes of real-world failure, this approach to fault modelling has not yet been examined in detail in work towards SRS fault detection. 
    \item The susceptibility of all robots in the SRS to faults and failures is a novel challenge for fault detection that previous work, in which the majority of the SRS typically retain uniform normal functionality, has not had to address.

\end{itemize}

The rest of this paper is structured as follows:
Section two describes the methodology and experimental test-bed used for this work. Section three details
the experiments performed to test the performance of the AAPD Model, including results and discussion. Section four concludes and lists avenues for future work.

\section{Methods}
To assist the reader, a list of all symbols used in this paper is provided below, along with their definitions.

\begin{itemize}
    \item $\mathcal{N}$: Gaussian noise about mean $\mu$ with standard deviation $\sigma$. $\mu = 0$ in all cases. $\sigma$ is always 5\% of the value to which it is added -- i.e. for a robot with its left wheel set to maximum velocity output, $v_l = v_{max} + \mathcal{N}(0,\frac{5}{100} v_{max})$. 
    \item $N$: The number of robots in a SRS. $1 \leq N \leq 20$.
    \item $R_{1-N}$: A robot within a SRS of size $N$.   
    \item $r$: Robot sensing range. See \autoref{equ:range_2}.
    \item $r_{max}$: Maximum robot sensing range, $r_{max} = 4m$.
    \item $P_{max}$: Maximum robot power capacity. Set to $P_{max} = 1$ or $P_{max} = \infty$.
    \item $v_{max}$: Maximum linear velocity of robot wheel, $0.22m s^{-1}$.
    \item $v_{l,r}$: Linear velocity of left and right robot wheels, respectively. See \autoref{equ:velocity}.
    \item $v$: Linear velocity of robot. $v = \frac{1}{2}(v_l + v_r)$.
    \item $a$: Axial separation of left and right robot wheels. $a = 16cm$.
    \item $\omega$: Angular velocity of robot. $\omega = \frac{1}{a}(v_r - v_l)$.
    \item $\Delta P_{max}$: Maximum rate of power consumption by robot. $\Delta P_{max} = \frac{1}{300} s^{-1}$.
    \item $\Delta P$: Rate of power consumption by robot. $\Delta P = \Delta P_s +\Delta P_{l} +\Delta P_{r}$.
    \item $\Delta P_{{l_{max}},{r_{max}}}$: Rate of power consumption by left and right motors at maximum load, respectively. $\Delta P_{{l_{max}},{r_{max}}} = \frac{2}{5}\Delta P_{max}$.
    \item $\Delta P_{l,r}$: Rate of power consumption by left and right motors, respectively. See \autoref{equ:power_wheels}.
    \item $\Delta P_S$: Rate of power consumption by sensing and communication. See \autoref{equ:P_S}.
    \item $d_{l,r}$: Degradation severity coefficient on left and right wheels, respectively. $0 \leq d_{l,r} \leq 1$.
    \item $d_S$: Degradation severity coefficient on localising signal transmitter. $0 \leq d_S \leq 1$.
    \item $d_0$: The ideal value of $d_{l,r,s}$ at which to detect a robot as faulty. $d_0\approx0.75$ (\autoref{fig:5}).
    \item $X$: The repertoire of all artificial antibody populations for a single robot. $X_{M,S}$ specifies motor or sensor hardware, respectively.
    
    \item $Y$: A learned repertoire containing paratopes labelled as faulty. $Y_{M,S}$ specifies motor or sensor hardware, respectively.
    
    \item $x_i$: The $i^{th}$ antibody population contained in $X$. See \autoref{equ:ok}.
    \item $F$: The fault threshold whereby an artificial antibody population is detected as faulty if $x_i > F$. $F = 1$.
    \item $p_i$: The paratope of an artificial antibody population $x_i$.
    \item $\hat{p}_j$: The $j^{th}$ labelled paratope in $Y$.
    \item $l$: The number of data entries encoded in a paratope dimension. $l = 30$.
    \item $W$: The recording of robot sensor and state data over a recent temporal window. $|W| = 300$.   
    \item $m$: The matching specificity between two paratopes. See \autoref{equ:ok_spec}.
    \item $s$: A threshold to enforce a minimum matching specificity between two paratopes. See \autoref{equ:ok_spec}.
    \item $\delta$: The value of $d_{l,r}$, whichever is smallest, at the moment an artificial antibody population is stimulated to $x_i > F$ where $x_i \in X_M$, or the value of $d_S$ at the moment an artificial antibody population is stimulated to $x_i > F$ where $x_i \in X_S$.
    \item $\Psi_T$: The time in which a robot is correctly detected as faulty as a proportion of the total experimental time it spends in a faulty state.
    \item $\Psi_F$: The time in which a robot is incorrectly detected as faulty as a proportion of the total experimental time it spends in a non-faulty state.

\end{itemize}

\subsection*{The Artificial Antibody Population Dynamics Model}

The AAPD model presented in this work is based on differential drive mobile robots with the ability to localise and estimate the relative positions of neighbouring robots, as well as send and receive data over a wireless network. This makes the proposed system suitable for implementation on most of the swarm robot platforms used in previous fault tolerant SRS research -- e.g. e-puck \cite{mondada2009puck}, marXbot \cite{bonani2010marxbot}, psi-swarm \cite{hilder2016psi}, or turtlebot3 \cite{TB3} platforms.

There are two key features of the natural immune system that are desirable to capture in autonomous fault tolerance systems. One is the immune systems ability to remember and more effectively combat familiar infection on subsequent encounter. The other is the immune system's ability to tolerate its host's own cells.

The natural immune system comprises an exceedingly large repertoire of lymphocytes, a type of white blood cell, that recognise and combat infection \cite{owen2013kuby}. Recognition is achieved using antibodies, proteins which are produced by lymphocytes, that bind to other cells according to their shape -- known as `lock and key' binding. When the immune system encounters foreign infection, the cells that recognise pathogens proliferate and differentiate \cite{owen2013kuby}. Once the infectious cells are destroyed, the immune cells which most effectively fought the infection are retained as memory cells \cite{kindt2007kuby}. These cells enable a faster and more efficient immune response should the host encounter the same infection again \cite{floreano2008bio}. If the immune system conflates infectious and domestic cells, it can potentially kill its host -- known as \textit{autoimmunity} \cite{langman2000third}. Immune network theory \cite{jerne1974towards} supposes that any immune cell can itself be recognised by a sub-section of the total immune cell repertoire, and that it is these domestic interactions that occur in the absence of any infectious cell that result in self-tolerance as an emergent property of the immune system. 

Farmer, Packard,and Perelson \cite{farmer1986immune} use immune network theory as the basis for their simplified immune model of antibody population dynamics. The model proposes that, when an antibody, the host's own cell, recognises and binds to another antibody or an antigen, an infectious foreign cell, the concentration of the binding antibody increases in proportion to the strength of recognition. When an antibody is recognised and bound to by another antibody, the concentration of the bound antibody decreases in proportion to the strength of binding. Antibody repertoire diversity and self-tolerant equilibrium is achieved through mutual self-interactions. Applying the models in \cite{farmer1986immune} to autonomous fault detection, artificial `antibodies' are encoded as small temporal windows of robot behaviour.

The AAPD model for a SRS can be described as follows. Robot state and sensor data are sampled at frequency $f_s$ during operation. Artifical antibody populations encode $l$ entries of one or more indicative metrics, such as linear or angular velocity, corresponding to a recording of robot behaviour over a window of time equal to $\frac{l}{f_s}$. This data comprises the signature, or `paratope', $p$, of an artificial antibody population, $x$. Each time a new artificial antibody paratope, $p_i$, is created (i.e. after each passing of $\frac{l}{f_s}$ seconds), it is assigned to an artificial antibody population $x_i$ that is initialised to zero and either added to a robot's artificial antibody repertoire, $X$, or discarded in an iterative process. A new artificial antibody population with paratope $p_i$ is only added to repertoire $X$ if $X$ does not already contain an artificial antibody population with a similar paratope $p_j$ such that $m(p_i,p_j) \leq u$, where $m$ is the matching specificity between two paratopes and $u$ is a matching threshold. 

The matching specificity $m(p_i,p_j)$ between the paratopes $p_i$ and $p_j$ is obtained by summing the residuals of the two paratopes as they are convolved over one another and averaging. This process is described by \autoref{equ:ok_spec}.

\begin{equation}
m(p_i,p_j) = \frac{1}{dim} \sum_{dim} \frac{1}{|\kappa|} \sum_{\in \kappa} G \big[s - \sum^{\eta}_{n} \big[ p_i(n) - p_j(n) \big] \big]
\label{equ:ok_spec}
\end{equation}

Where $dim$ is the number of dimensions used to construct the paratopes $p_{i,j}$. $\eta$ is the number of overlaying data points between two paratopes, $p_{i,j}$, of sizes $l_{i,j}$. $n$ is the index of data points in the overlapping range $\eta$. $k$ is the maximum number of allowable datapoints that can be without partner as two paratopes are convolved over one another. $\kappa$ is the set of all points of convolution such that $\in k = 0:g:\tau$ where $\tau = l_i - l_j + k + 1$. For $k=0$, two paratopes of equal size will only have one possible convolution point from which a residual can be taken -- e.g. $\tau = 1$. For mismatched paratope sizes, the number of possible points of convolution is determined by the difference in size and the maximum allowable offset $k$. The value $g$ can be set to $g = 1$ to give a comprehensive convolution across every possible index, or increased to give a sparser convolution at decreased computational cost. Terms $G$ and $s$ are used to enforce a matching threshold. $G(x) = x$ for $x > 0$, and $G(x) = 0$ otherwise. As the sum of residuals between $p_i$ and $p_j$ approaches 0, $G(x)$ approaches $s$. If the summed residuals are greater than $s$, matching is considered insufficient to be counted and is discarded. \autoref{equ:ok_spec} is functionally parallel with the equation for matching specificity used by Farmer et al. \cite{farmer1986immune}.

Each robot in the SRS independently computes the dynamics of the artificial antibody populations in its own repertoire, $X$, by comparing the paratope of each population with its own recent behavioural history and with that of its neighbours. In addition to writing relevant state and sensor data to paratopes, each robot writes the same data to a rolling behavioural window, $W$, at frequency $f_s$. Unlike paratopes, which, once written and added, can remain in repertoire $X$ indefinitely, behvioural window $W$ always contains the most recent $\frac{|W|}{f_s}$ seconds worth of robot data. 

Once enough time has passed for each robot to have filled behavioural window $W$ to capacity, one or more artificial antibody populations, $x$, will have then been added to repertoire $X$, and the AAPD model computes the population dynamics for each member of $X$ according to \autoref{equ:ok}. 

\begin{equation}
\dot{x_i} =  m(p_i, W_{self})\cdot(1 + k_1 \cdot max [ m(p_i, Y)]) - k_2 \sum_{j=1}^{N-1} m(p_i, W_j) - k_3 
\label{equ:ok}
\end{equation}

Where $x_i$ is the population level of an artificial antibody with paratope $p_i$. $W_{self}$ is the robots own behavioural window and $W_j$ is the behavioural window of the $j^{th}$ neighbour in the swarm. $Y$ is a repertoire containing paratopes that are labelled as faulty, either because they have been previously detected by the AAPD model or they are taken from training data. $k_{1,2,3}$ are tuning coefficients. 

\autoref{equ:ok} can be described thus: For a SRS of $N$ robots, an artificial antibody population, $x_i$, in a given robot's repertoire $X$ is stimulated according to the matching specificity between its paratope $p_i$ and the robot's own behavioural array, $W_{self}$, according to \autoref{equ:ok_spec}. This has the effect of providing greater stimulation to artificial antibody populations with paratopes that have been frequently exhibited in $W$. $x_i$ is further stimulated if $Y$ contains a paratope $p_j$ that closely matches $p_i$. Where paratope $p_i$ matches with multiple paratopes in $Y$, the population of $x_i$ is stimulated according to the paratope in $Y$ that produces the greatest matching specificity with $p_i$. At the same time, artificial antibody population $x_i$ is suppressed according to the matching specificity between paratope $p_i$ and the behavioural arrays of the other robots in the swarm, $W_{1 - (N-1)}$.

Each robot in the SRS runs its own instance of the AAPD model that computes once the robot's behavioural array $W$ is full, and then recurrently once enough time has passed such that each value of $W$ used in the previous computation has been replaced by a new value. At each computation of the AAPD model, if artificial antibody population $x_i$ drops below zero, it is removed from repertoire $X$. If the population of artificial antibody $x_i$ is stimulated above fault threshold $F$, such that $x_i > f$, it is treated as indicative of a fault in the robot to which it belongs. Each robot maintains its own independent repertoire $X$. Repertoire $Y$ is shared across the swarm with each new addition so that all robots benefit from the learning process.

\begin{figure*}[!t]
  \centering  
    \includegraphics[width=\textwidth]{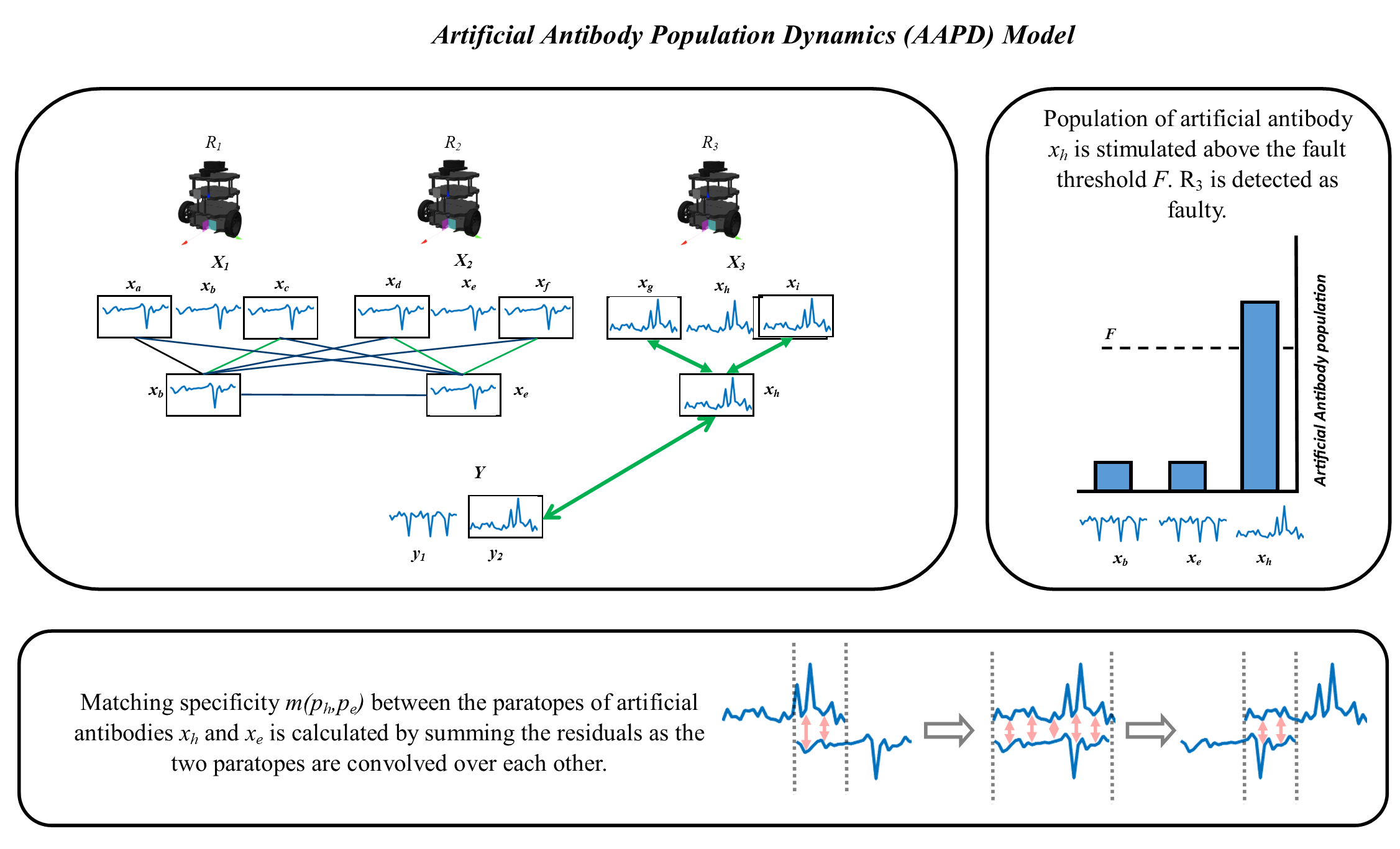}
    \caption{The AAPD-model runs on a TurtleBot3 SRS (robots $R_{1-3}$ shown). Temporal samples of
robot state and sensor data (linear velocity shown) are encoded in artificial antibodies, for which each robot has it's own repertoire (${X}_{1-3}$). $R_1$’s artificial antibody $x_b$ has a high matching specificity, $m$, with antibodies $x_a$ and $x_c$ according to \autoref{equ:ok_spec}, resulting in stimulation of population $x_b$. The same is true for $R_2$'s
antibody $x_e$ with $x_d$ and $x_f$. The high matching specificity between the artificial antibodies of $R_1$ and $R_2$ ($x_b$ with $x_d$, $x_e$,
and $x_f$, and $x_e$ with $x_a$, $x_b$, and $x_c$) results in the mutual suppression of populations of $x_b$ and $x_e$ such that they are tolerated as normal. $R_3$’s artificial antibody $x_h$ has a high matching specificity with artificial antibodies $x_g$ and $x_i$,
resulting in population stimulation of $x_h$, however $x_h$ has a low matching specificity with the
antibodies of $R_1$ and $R_2$ because of the high residuals when they are convolved with \autoref{equ:ok_spec}, meaning that the population of $x_h$ is not suppressed. $x_h$ is further stimulated by its high matching specificity with paratope $y_2$, contained in $Y$, which further stimulates the population of $x_h$, taking it
over the threshold $f$ for detection of $R_3$ as faulty. }  
 \label{fig:1}
\end{figure*}

A simplified illustration of the AAPD model can be seen in \autoref{fig:1}.

\subsection*{Experimental Test-Bed}

This work considers a simulated SRS of TurtleBot3 robots \cite{TB3} (shown in \autoref{fig:1}). The TurtleBot3 is a two-wheeled differential drive robot with open source models for simulated experimentation in ROS and Gazebo. Simulated robots are provided with the ability to communicate wirelessly with each other over a simulated network, and to physically detect and locate other robots and objects up to a maximum distance of 4 metres away. 

The SRS performs an autonomous foraging task, a common benchmark in swarm research \cite{bayindir2016review}, in an enclosed arena measuring 10m by 10m. Two types of foraging algorithm are considered. In the global positioning foraging (GPF) algorithm, described by \autoref{alg_fo}, each robot must retrieve a resource from one of three circular resource nests with a radius of 1m, distributed evenly at arena ($x,y$) coordinates (2,8), (5,8), and (8,8), respectively. The resource must then be returned to an area referred to as the `base', which spans the entire $x$ dimension of the arena for $0 \leq y \leq 2$. Robots are assumed to have access to global positioning data for localisation. The Local Positioning Foraging (LPF) algorithm, described by \autoref{alg_fo_2}, is similar to the GPF algorithm, except that robots are not assumed to have access to GPS information. In order to localise, the swarm must form an ad-hoc network. Each robot has a limited localising range, and its status is determined by whether or not there exists a path from a robot to the base that is valid for the variable sensing ranges of each node. A robot is only able to extend a communication chain if it is within the sensing range of the previous node. A robot will not move if it cannot localise.

\begin{figure}[!t]
  \centering  
    \includegraphics[width=\textwidth]{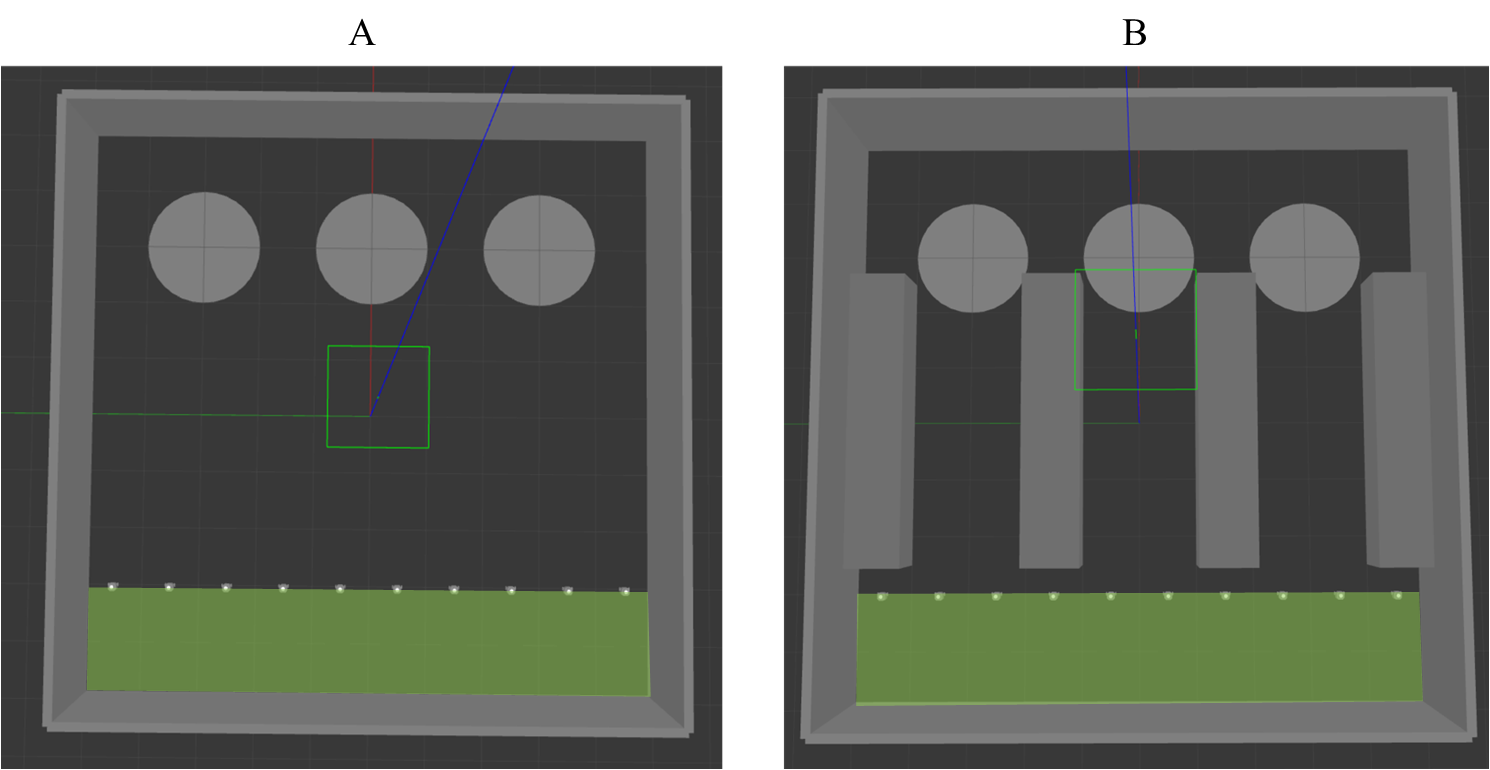}
    \caption{Experimental setup for 10 robots performing \autoref{alg_fo} or \autoref{alg_fo_2} in an enclosed empty environment (A) or constrained environment (B). Resource nests are indicated by the three grey circles opposite the robots. The highlighted green area indicates the robot base.} 
 \label{fig:2}
\end{figure}

Two types of arena are considered. The first, referred to as the `empty' arena, is an enclosed 10m x 10m arena containing the three resource nests described previously with no other obstacles except for the robots operating within it. This can be seen in \autoref{fig:2}A. The second, referred to as the 'constrained' arena, is also an enclosed 10m x 10m arena containing 3 resource nests at the same positions. However, the area between the resource nests and the robot base is separated into 3 equally spaced corridors of 2m width and 5m length. This can be seen in \autoref{fig:2}B.

\begin{algorithm} [t]
\caption{Foraging Algorithm}\label{alg_fo}
\begin{algorithmic}[1]
\While {Running}
\If {Object Distance $\leq 0.5m$} avoid
\ElsIf{Resource collected \textbf{or} Battery low \textbf{or} Faulty} Return to base
\If{Robot at base} Deposit resource
\EndIf
\If{Battery low} Recharge
\EndIf
\If{Faulty} Repair
\EndIf
\ElsIf {Distance to centre of nearest Resource Nest $\leq 0.75m$} Collect resource
\ElsIf {Distance to nearest Resource Nest $\leq r$} Approach nearest resource
\Else{} Move forwards
\EndIf
\EndWhile
\end{algorithmic}
\end{algorithm}

\begin{algorithm} [t]
\caption{Local Positioning Foraging (LPF) Algorithm}\label{alg_fo_2}
\begin{algorithmic}[1]
\While {Running}
\If{Distance to closest networked node $\leq 3m$}
\If {Object Distance $\leq 0.5m$} avoid
\ElsIf{Resource collected} Return to base
\If{Robot at base} Deposit resource
\EndIf
\ElsIf {Distance to nearest Resource Nest $\leq 0.5m$} Collect resource
\ElsIf {Distance to nearest Resource Nest $\leq r_s$} Approach nearest resource
\Else{} Random Explore
\EndIf
\Else{} Wait
\EndIf
\EndWhile
\end{algorithmic}
\end{algorithm}

\subsubsection*{Electro-Mechanical Fault Modelling}

The sensing, communication, and locomotion functions performed by each robot consume power and are affected by degradation at different rates. A proportional model of power consumption is utilised whereby each robot is initialised with maximum power $P_{max} = 1$ and each process consumes a percentage of the robot's maximum power output, $\Delta P_{max}$, per unit time. It is expected that locomotion consumes significantly more power than sensing, communication, and other background processes. Power consumption for a robot with both motors drawing maximum load, the most power consuming state it can take, is modelled as a 20:40:40 split between power consumed by sensing, communication, and other background processes, the power consumed by the left motor, and the power consumed by the right motor, respectively. $\Delta P_{max} = \frac{1}{300}$ so that a robot drawing maximum power can operate uninterrupted for a total of five minutes of simulated time.

Performance reductions caused by gradual wear, the accumulation of dust and debris, and adverse environmental conditions is simulated via a degradation severity coefficient $d_{l,r}$ on left and right motors, respectively, and $d_S$ on the transmission range of simulated sensors. Degradation affects physical processes and the power consumed by them. $d_{l,r,S}$ can take any value between 0 and 1, with 1 indicating perfect condition and 0 indicating that the corresponding hardware is completely degraded. The power consumption and degradation models for sensing, communication, and locomotion are as follows.

The power consumed by robot locomotion will be affected by the condition of its wheels and motors. Motors are typically designed to run at 50\% - 100\% of rated load, with maximum efficiency at around 75\% \cite{ranges26determining}. A robot with $d_{l,r} = 1$ is therefore taken to cause its motors to operate at 75\% load.

The mechanical power required to achieve a given robot velocity is proportional to the forces incident upon it. As motor hardware degrades, resistance increases along with the mechanical power required to maintain a given velocity. A study into the reliability of motors over time plots the relationship as an approximately sigmoidal function \cite{zhang2022performance}. Since the reliability and efficiency of a motor are directly related \cite{ranges26determining}, the effects of motor degradation on robot locomotion are modelled as follows.

The rate of power consumption, $\Delta P_{l,r}$, by left and right motors, respectively, is given by \autoref{equ:power_wheels}.

\begin{equation}
     \Delta P_{l,r} = \frac{\Delta P_{{l_{max}},{r_{max}}}}{1+e^{-10((1-d_{l,r}) + 0.11)}} + \mathcal{N}   
     \label{equ:power_wheels}
\end{equation}

where values of constants are set such that $P_{l,r} \approx 0.75 \Delta P_{{l_{max}},{r_{max}}}$ for $d_{l,r} = 1$. The standard deviation of Gaussian noise applied to $\Delta P_{l,r}$ is set to be 5\% of the base value of $\Delta P_{l,r}$. 

Robot velocity is modelled as:

\begin{equation}
    v_{l,r} = \frac{v_{max}}{1+e^{-5(2d_{l,r} - 1)}} + \mathcal{N}
    \label{equ:velocity}
\end{equation}

The standard deviation of Gaussian noise applied to $\Delta P_{l,r}$ is set to be 5\% of the base value of $\Delta P_{l,r}$. 

\begin{figure*}[!t]
    \includegraphics[width=\textwidth]{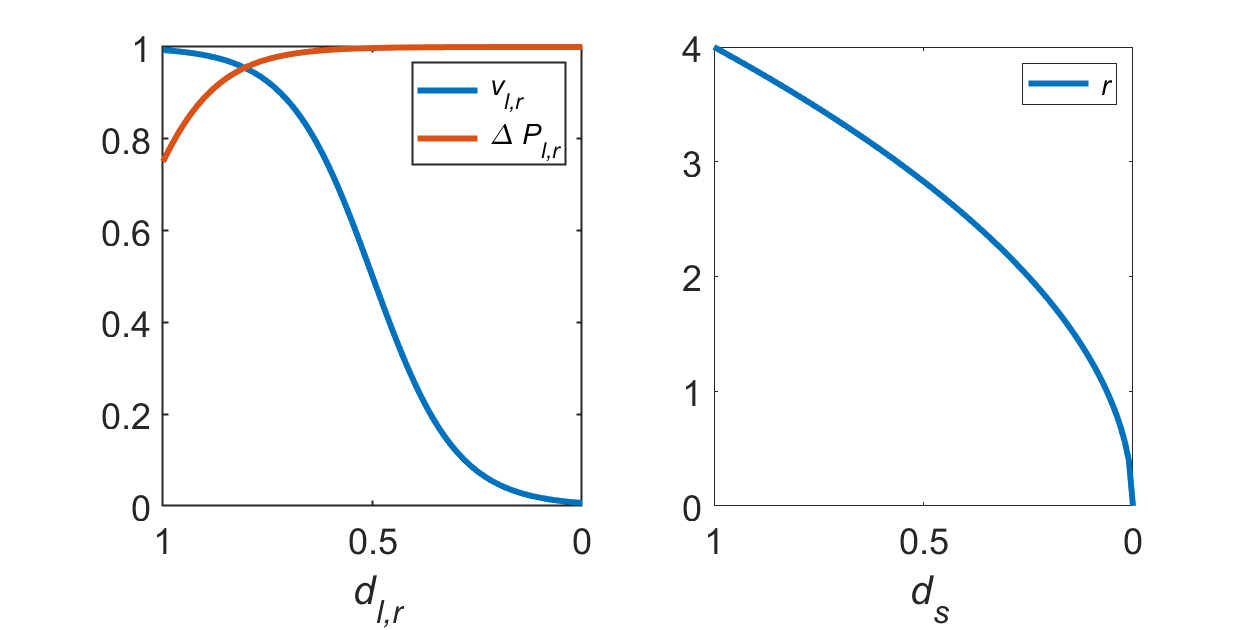}
    \caption{Velocity and power consumption of left or right wheels, $v_{l,r}$ and $\Delta P_{l,r}$, respectively, normalised and plotted against degradation severity coefficients $d_{l,r}$ according to \autoref{equ:velocity} and \autoref{equ:power_wheels} (left plot). Robot sensing range $r$ plotted against degradation severity coefficient $d_S$ according to \autoref{equ:range_2} (right plot). }  
 \label{fig:3}
\end{figure*}

The values of constants in \autoref{equ:power_wheels} and \autoref{equ:velocity} are set to give the intersection of $v_{l,r}$ and $\Delta P_{l,r}$ as plotted in \autoref{fig:3}. This intersection reflects that, as the value of $d_{l,r}$
increases, motors can initially draw more power to achieve $v_{max}$ but, eventually, the mechanical power required will become greater than that which can be supplied by the robot battery. At this point, degradation will begin to reduce a robots maximum achievable velocity.

Each robot is constantly emitting data and receiving data from other robots. Since transceiver outputs are not governed by corrective feedback loops in this case, the power consumed by robot sensing hardware and background processes, $\Delta P_S$, is not expected to vary outside of a normal range. $\Delta P_S$ is therefore modeled as a constant with added Gaussian noise.

\begin{equation}
    \Delta P_S = \frac{1}{5}\Delta P_{max} + \mathcal{N}
    \label{equ:P_S}
\end{equation}. 

The standard deviation of Gaussian noise applied to $\Delta P_S$ is set to be 5\% of the base value of $\Delta P_S$. 

Degradation on the robot sensor, or environmental variations that result in higher levels of signal attenuation, will affect the sensing range, $r$. This is modelled according to the inverse square law in \autoref{equ:range_2} and plotted in \autoref{fig:3}.

\begin{equation}
    r = r_{max} \sqrt{d_s} + \mathcal{N}
    \label{equ:range_2}
\end{equation}

The standard deviation of Gaussian noise applied to $r$ is set to be 5\% of the base value of $r$.

\subsection*{Implementation of the AAPD Model on SRS}

Deploying the AAPD model on a SRS requires that robot behaviour is mapped to the paratopes of artificial antibody populations. Degraded motor hardware produces readily and endogenously measurable effects in each robot according to the values of $d_{l,r}$ so long as the robot is moving, affecting robot linear velocity, $v$, angular velocity, $\omega$, and power consumption, $\Delta P$. 

Sensor degradation affects transmission range, $r$, which cannot be measured endogenously. Therefore, a type of handshake protocol is used, for which a new matric, $\gamma$, is introduced. $\gamma$ indicates the closest distance at which a given robot $R_1$ can sense neighbour $R_2$, but where $R_2$ is simultaneously unable to sense $R_1$. $\gamma = r_{max}$ if a robot can mutually sense and be sensed by each neighbour in its own sensing range $r$. 

\begin{figure}[!t]
  \centering  
    \includegraphics[width=0.8\textwidth]{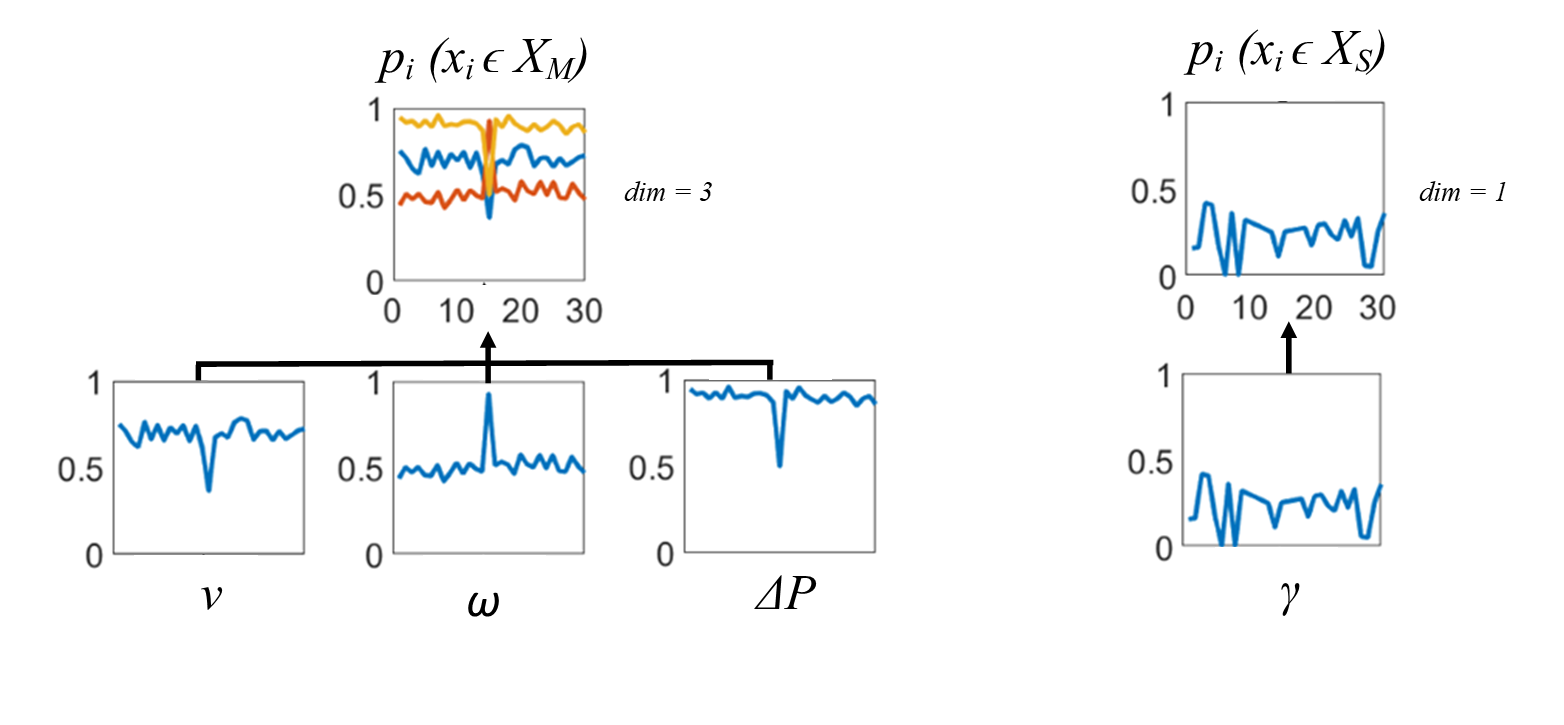}
    \caption{The construction of artificial antibody paratopes for robot motors, $p_M$, and sensors, $p_S$. $p_M$ has a 3 dimensional paratope consisting of 5 second recordings ($l = 30$) of robot linear velocity, $v$, angular velocity, $\omega$, and rate of power consumption, $\Delta P$. $p_S$ has a 1 dimensional paratope consisting of a 5 second recording of $\gamma$.}  
 \label{fig:4}
\end{figure}

The shapes of $v, \omega$ and $\Delta P$, or $\gamma$, plotted to a normalised common axis over time, are thus used to represent the paratopes of artificial antibody populations. Artificial antibody populations corresponding to motor hardware, $x_M$, and sensor hardware, $x_S$, are written to separate repertoires, $X_M$ and $X_S$, and handled separately by the AAPD model (see \autoref{fig:4}). 

Each paratope encodes five seconds of $v, \omega$ and $\Delta P$, or $\gamma$ data sampled at $f_s = 6Hz$ -- i.e. $l = 30$. Behavioural data windows, $W_{M,S}$, contain the most recent fifty seconds of $v, \omega$ and $\Delta P$, or $\gamma$ data, respectively, sampled at the same rate -- i.e. $|W| = 300$.

Repertoire $Y$ contains paratopes that have been labelled as faulty. This work implements an unsupervised learning process whereby $Y$ contains paratopes that have been detected by the AAPD model itself. In order to standardise this process, AAPD model `orders' are introduced. To illustrate, an instance of the AAPD model that is provided with an empty repertoire $Y$ can be considered a zeroth order model. A zeroth order model functions as a data-driven outlier detection model, comparable to the models presented in \cite{taraporePLOS, carminati2024distributed}, and represents the most basic implementation of the AAPD model. When faults are detected by the zeroth order model, the paratopes of artificial antibody populations for which $x > F$ are added to $Y$. As with the addition of artificial antibody populations to repertoire $X$, a new paratope $p_i$ is only added to repertoire $Y$ if $Y$ does not already contain a paratope $\hat{p}_j$ such that $m(p_i,\hat{p}_j) \leq u$. A repertoire $Y$ that contains paratopes labelled as faulty by a zeroth order AAPD model is then considered a first order repertoire, $Y_1$. A first order AAPD model is provided with a first order repertoire, and any paratopes as faulty are then added to a second order repertoire, $Y_2$, according to the same process. A model provided with a second order repertoire is then considered to be a second order model, and so on. In this work, the repertoire $Y$ that can be accessed by an AAPD model never changes during an experiment. 

\subsection*{Problem Definition and Parameter Tuning}

The problems that this work seeks to address can now be articulated as follows. Given a SRS performing a foraging task while individual robot sensor and motor hardware are subjected to varying degrees of performance degradation:

\begin{enumerate}
    \item What is the ideal level of degradation, $d_0$, above which a robot should be tolerated and below which a robot be detected as faulty?
    \item Can AAPD model paramaters be tuned such that they can reliably detect robots with any value $d_{l,r,S} \leq d_0$ as faulty while tolerating robots with all values $d_{l,r,S} > d_0$?
    \item Can the paratopes of artificial antibody populations for which $x > F$ be used to diagnose faults?
    \item Can deployment of the tuned AAPD model prevent instances of robot failure in the field?
\end{enumerate}

\begin{figure}[!t]
  \centering  
    \includegraphics[width=0.8\textwidth]{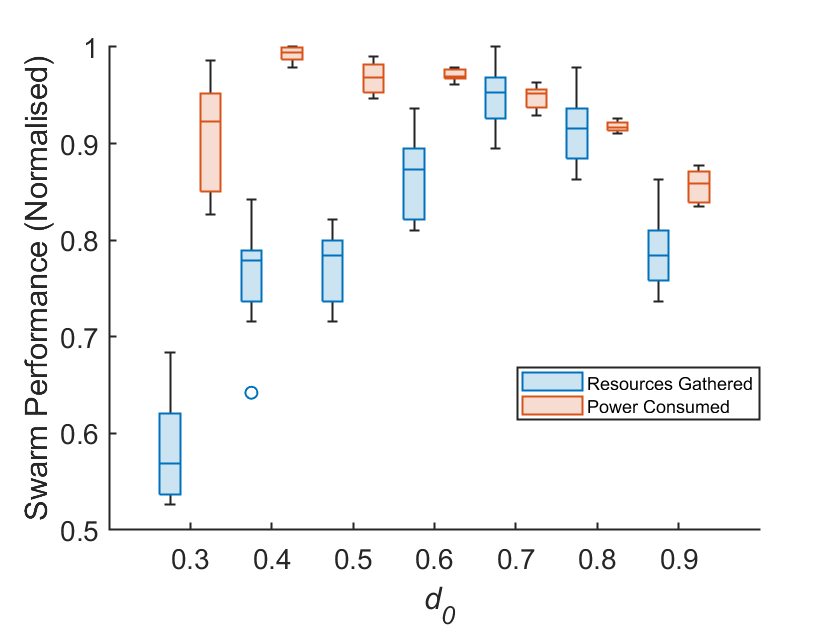}
    \caption{The resources collected and power consumed in 15 minutes by a SRS of $N = 10$ robots performing \autoref{alg_fo}. Maintenance is scheduled when a robot has any value $d_{l,r,S} < d_0$. Data presented is normalised to a common y axis.}  
 \label{fig:5}
\end{figure}

Addressing these problems, which are detailed in the following section, first requires that the value $d_0$ be found and AAPD model parameters be tuned. To identify the ideal value of $d_0$, a series of preliminary experiments are conducted with a SRS of $N = 10$ robots performing the GPF algorithm in the open environment. Robots are initialised evenly and horizontally about the midpoint (5,2), as seen in \autoref{fig:2}. Each robot is initialised with $P = 1$ and random independent probabilities, between 1 - 15\%, that their respective values $d_l$, $d_r$, and $d_S$ will decrement by 0.01 per second of simulated time. When any of the degradation coefficients of a given robot drops below threshold $d_0$, that robot is declared as faulty and returns itself to the base. Robots must return to base in order to receive charge or maintenance work. If a robot depletes its power or degrades to a point of immobility outside of the base, it is considered lost and becomes an obstacle for the remainder of an experiment. If the robot successfully returns to the base, it is assumed to receive the required maintenance or power replenishment before being redeployed. The total number of resources collected and power consumed by the swarm in 15 minutes of simulated time are monitored. By varying the value $d_{0}$, an appropriate point for maintenance scheduling can be found. The results across 10 experimental replicates are plotted in \autoref{fig:5}.

\autoref{fig:5} shows that setting $d_0$ in the range $0.7 \leq d \leq 0.8$ produces the best performance in terms of total resources collected in the scenario tested. $d_0 = 0.7$ offers marginally better performance in terms of total resources collected at the expense of a slight increase in the total power consumed. Scheduling maintenance at $d_0 > 0.8$ results in more frequent interruptions to normal operation, while the physical effects of degradation are more obstructive at $d_0 < 0.7$. For $d_0 \leq 0.3$, some robots become stuck, unable to complete the return journey to base until, eventually, their power is depleted. This is why there is a sharp decrease in the total resources collected. Unsurprisingly, the system generally consumes more power the longer it is left to degrade. The exception to this trend is where $d_0 \leq 0.3$ -- caused by robots that deplete their power outside of the base being thus unable to continue consuming power. In general, it can be seen that the power consumed by the SRS does not change much for $0.4 \leq d_0 \leq 0.8$, and so cannot give a strong indication of SRS performance in this range. It is therefore omitted from consideration in the parameter tuning process and subsequent performance assessments. $d_0 = 0.75$ is selected as the target for autonomous detection of degradation and maintenance scheduling.

Parameters of the AAPD model can now be found such that artificial antibody populations produced by robots with any value $d_{l,r,S} \leq 0.75$ are stimulated above threshold $F$ as quickly as possible, while the populations of artificial antibodies produced by robots with all values $d_{l,r,S} > 0.75$ are kept below threshold $F$.

Training data is taken from a set of experiments (10 replicates) whereby a SRS of $N = 10$ robots performs the GPF algorithm in the open environment for 15 minutes of simulated time. All robots are initialised with $P = \infty$, since robots that fully deplete their power are not useful for training model parameters. Robot $R_1$ is initialised with a probability between 5-15\% of $d_{l,r,S}$ decrementing by 0.01 per second of simulated time while robots $R_{2-10}$ are initialised with $0.75 < d_{l,r,S} < 1$. The $v, \omega, \Delta P$ and $\gamma$ data from each robot is sampled at 6Hz and recorded for the entire experiment duration. This data is then used to train AAPD model parameters offline. Parameters are tuned for the zeroth, first, and second AAPD models, since improved performance was not observed for model orders greater than two. For each unique set of parameters tested on any given experimental replicate, the first and second order AAPD models are provided with repertoire $Y$ consisting of paratopes labelled as faulty by the zeroth and first order models with the same parameters, respectively, compiled across all experimental replicates except for the current test index. This is done to prevent $Y$ containing identical paratopes with those emerging from the current test data, as this was observed to interfere with the training process. 

The performance of the AAPD model is assessed on the following criteria:
\begin{itemize}
    \item $\delta$: The value of $d_{l,r}$, whichever is smallest, at the moment a motor fault is detected by the AAPD model or the value of $d_S$ at the moment a sensor fault is detected by the AAPD model.
    \item $\Psi_T$: True positive detection rate, or the experimental time during which a robot has any artificial antibody population detected as faulty by the AAPD model while it has one or more degradation severity coefficients $d_{l,r,S} \leq 0.75$ as a proportion of the total experimental time it spends with one or more degradation severity coefficients $d_{l,r,S} \leq 0.75$.
    \item $\Psi_F$: False positive detection rate, or the experimental time during which a robot has any artificial antibody population detected as faulty by the AAPD model while it has all degradation severity coefficients $d_{l,r,S} > 0.75$ as a proportion of the experimental total time it spends with all degradation severity coefficients $d_{l,r,S} > 0.75$.
\end{itemize}

Optimising the model parameters for these criteria was surprisingly challenging - early efforts to use gradient descent were unsatisfactory, mostly because of the difficulty in applying $\delta$ and $\Psi_{T,F}$ as quantitative fitness measures to gradually degrading robots. In practice, there is not enough behavioural distinction between, for example, a robot with $d_l = 0.74$ and a robot with $d_l = 0.76$ to achieve reliable true positive fault detection in this region -- especially if $d_r$ remains high. Setting any instance of fault detection where $d_{l,r,S} > 0.75$ as a false positive instance and punishing those parameter combinations during the gradient descent process results in model parameters that struggle to detect true positive instances until $d_{l,r,S}$ are much lower than desired. Furthermore, detecting a fault for a robot where $d_{l,r,S} = 0.76$, a false-positive, is preferable to detecting at $d_{l,r,S} = 0.6$, a true-positive, if the ideal point of detection is $d_0 = 0.75$. This illustrates that the boundary between true and false positive detection is much more difficult to define for gradual degradation -- especially where multiple degradation coefficients contribute to overall behaviour, as with $d_l$ and $d_r$. There is also a distinction between model behaviours that should be actively punished and those which should merely be improved upon. For example, detecting a fault once a robot has some value $d_{l,r,S} \approx 0.3$ and sustaining the detection thereafter is not ideal. However, this is nonetheless a true positive detection that the AAPD model should be able to make reliably and should not therefore be punished by the learning process. Detecting a fault in a robot with some value $0.76 < d_{l,r,S} < 0.8$ is a false positive detection, but one that is arguably also undeserving of punishment by the learning process, since it falls within the ideal range shown in \autoref{fig:5}. If the AAPD model makes a false positive detection for a robot with, e.g., some value $d_{l,r,S} \approx 0.95$, but only momentarily, the punishment should also be less severe than if the detection was sustained. It is thus non-trivial to meaningfully quantify and offset the contributions of $\delta$ and $\Psi_{T,F}$ in the learning process without understanding the costs (e.g. time, energy expense, resources gained/lost, etc.) of robot operation, robot repair, and objective values. While this is highly relevant to the field of autonomous fault tolerance research, it requires domain-specific knowledge of robot platforms and use-case scenarios that is beyond the scope of this study. 
Parameter values were instead user-selected with the aim of maximising $\Psi_T$, minimising $\Psi_F$, and getting median average $\delta$ as close to 0.75 as possible. 
$k_{1,2,3}$ were tuned with a granularity of 0.01, while $u, F, s$ were tuned with a granularity of 0.1. This process revealed that different stages of the AAPD model performed better with different values of $s$ in \autoref{equ:ok_spec} and that parameter values of the AAPD model that performed well for repertoire $X_M$ did not necessarily perform comparably for repertoire $X_S$ and vice versa. 

Held constant throughout this work are $u = 1.2$ and $F = 1$. Variables $s, g, k$ in \autoref{equ:ok_spec} and $k_{1,3}$ in \autoref{equ:ok} vary according to what stage of the AAPD model they are applied to. These values are given as follows.

A new artificial antibody population $x_i$ with paratope $p_i$ is added to repertoire $X$ if $X$ contains no paratope $p_j$ such that $m(p_i,p_j) > u$. Similarly, an artificial antibody population for which $x_i > F$ has its paratope $p_i$ added to repertoire $Y$ if $Y$ contains no paratope $\hat{p_j}$ such that $m(p_i,\hat{p}_j) > u$. For each process, $m$ is computed with \autoref{equ:ok_spec} where $s = 1.5$, $g = 1$, and $k = 10$. This is true for paratopes relating to motor and sensor hardware.

Where $x_i \in X_M$, $\dot{x_i}$ is computed with \autoref{equ:ok} where $k_1 = 0.24$, $k_2 = 0.3$, and $k_3 = 1.2$. In this process, $m(p_i,W)$ is computed with \autoref{equ:ok_spec} where $s = 4$, $g = 5$, and $k = 0$. $m(p_i,\hat{p}_j)$ is computed with \autoref{equ:ok_spec} where $s = 1.5$, $g = 1$, and $k = 10$.

Where $x_i \in X_S$, $\dot{x_i}$ is computed with \autoref{equ:ok} where $k_1 = 0.18$, $k_2 = 0.3$, and $k_3 = 1.2$. In this process, $m(p_i,W)$ is computed with \autoref{equ:ok_spec} where $s = 5$, $g = 5$, and $k = 0$. $m(p_i,\hat{p}_j)$ is computed with \autoref{equ:ok_spec} where $s = 3.3$, $g = 1$, and $k = 10$.

For ease of reference, this information is also summarised in \autoref{table:param_values}.

\begin{table*}
\begin{center}
\begin{tabular}{ |l|c|c|c|c|c|c|c|}
\hline
Process & Computation & $s$ & $g$ & $k$ & $k_1$ & $k_2$ & $k_3$ \\
\hline
$x_i \cup X,Y$ & \autoref{equ:ok_spec} & 1.5 & 1 & 10 & - & - & - \\
$m(p_i,W)$ where $x_i \in X_M$ & \autoref{equ:ok_spec} & 4 & 5 & 0 & - & - & - \\
$m(p_i,\hat{p}_j)$ where $x_i \in X_M$ & \autoref{equ:ok_spec} & 1.5 & 1 & 10 & - & - & - \\
$\dot{x_i}$ where $x_i \in X_M$ & \autoref{equ:ok} & - & - & - & 0.24 & 0.3 & 1.2 \\
$m(p_i,W)$ where $x_i \in X_S$ & \autoref{equ:ok_spec} & 5 & 5 & 0 & - & - & - \\
$m(p_i,\hat{p}_j)$ where $x_i \in X_S$ & \autoref{equ:ok_spec} & 3.3 & 1 & 10 & - & - & - \\
$\dot{x_i}$ where $x_i \in X_S$ & \autoref{equ:ok} & - & - & - & 0.18 & 0.3 & 1.2 \\
\hline

\end{tabular}
\end{center}
\caption{The various parameter selections for \autoref{equ:ok} and \autoref{equ:ok_spec} used within different stages of the AAPD model}
\label{table:param_values}
\end{table*}

\section{Experiments, Results \& Discussion}
This section tests the performance of the AAPD model against the criteria described in the previous section, namely: $\delta$, $\Psi_{T,F}$, and the instances of failure in the field. A number of scenarios are considered, with ten experimental replicates performed for each set of experiments. All experiments are conducted with ROS 2 and Gazebo, with MATLAB (2022) used for offline processing.

\subsection*{Degradation in a Single Robot}

The first set of experiments tests the AAPD model's ability to detect faults in a single robot as it stochastically degrades from among a SRS exhibiting a range of tolerable $d_{l,r,S}$ values.
A SRS of $N = 10$ robots perform the GPF algorithm in the open environment. All robots are initialised with $P = \infty$. Robot $R_1$ is initialised with $0.9 < d_{l,r,S} < 1$ and independent random probabilities between 1-15\% of $d_{l,r,S}$ decrementing by 0.01 per second of simulated time. Robots $R_{2-10}$ are initialised with static $0.75 < d_{l,r,s} < 1$, $0.85 < d_{l,r,s} < 1$, or $0.95 < d_{l,r,s} < 1$. This is done in order to assess how the coverage of $d_{l,r,s}$ values in the non faulty range $d_{l,r,s} > 0.75$ in $R_{2-10}$ affects AAPD model performance. Separate sets of experiments are performed to examine motor and sensor degradation in isolation. The zeroth order model is used to produce a repertoire $Y$ to be used by the first order model, where $Y_{M.1,S.1}$ refers to the first order repertoire for motor and sensor paratopes, respectively. The first order model is then used to create second order repertoires $Y_{M.2,S.2}$.

$Y_{M.1}$ contains 99 unique paratopes, $Y_{S.1}$ contains 93 unique paratopes, $Y_{M.2}$ contains 101 unique paratopes, and $Y_{S.2}$ contains 109 unique paratopes. 

The performance of the AAPD model is assessed offline, with the model applied to the entire experimental recording of robot data after the experiment has ended. This allows for direct comparison of different model orders on the same sets of robot data.

\begin{figure}[!t]
    \includegraphics[width=\textwidth]{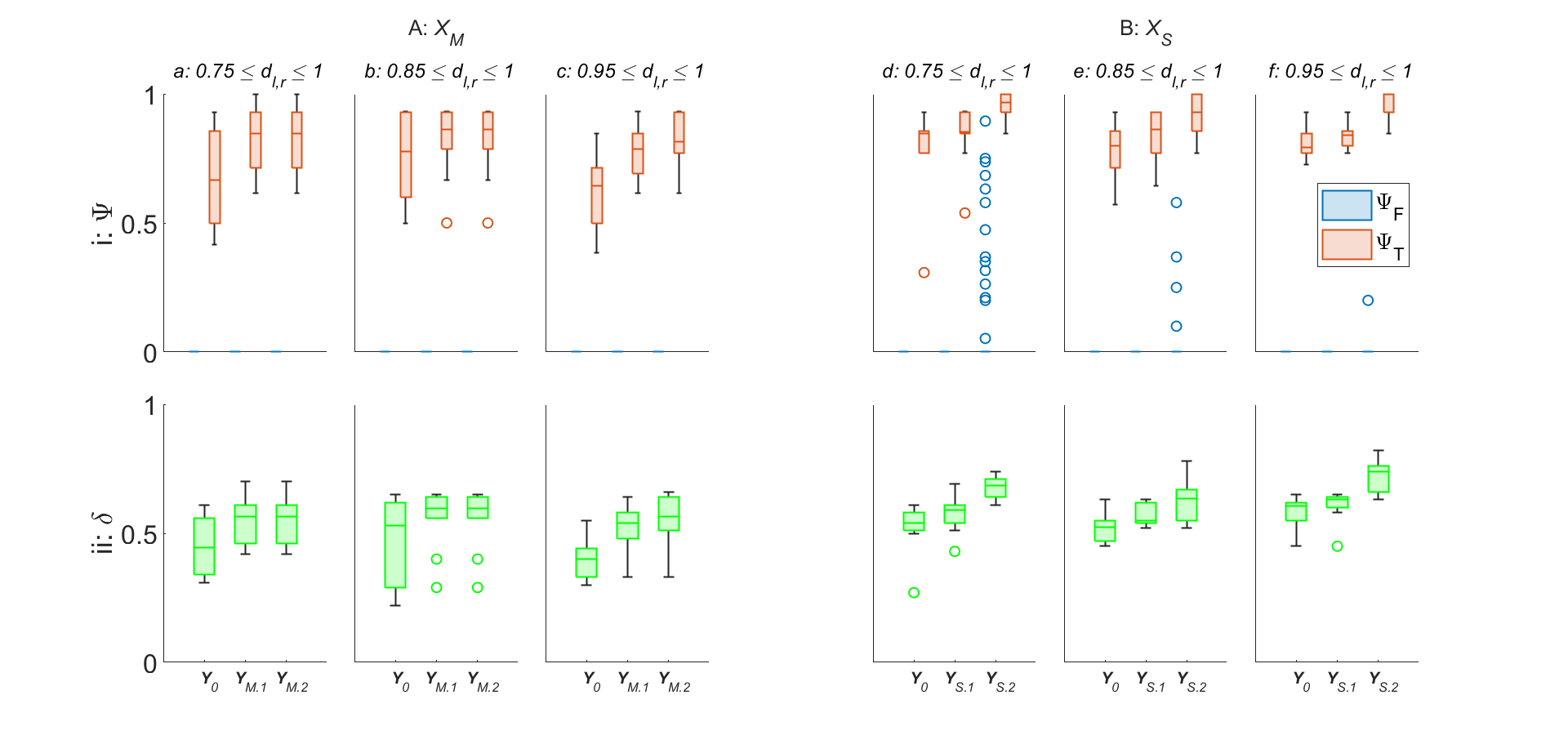}
    \caption{$\Psi_{T,F}$ (row i), and $\delta$ (row ii) for zeroth, first, and second order AAPD models operating on A: $X_M$ (columns a-c), denoted $Y_0$, $Y_{M.1}$, or $Y_{M.2}$, respectively, or operating on B: $X_S$ (columns d-f), denoted $Y_0$, $Y_{S.1}$, or $Y_{S.2}$, respectively, on a SRS of $N = 10$ robots. Columns a-f also indicate the range of $d_{l,r,S}$ values that non-faulty robots $R_{2-N}$ are initialised with.}  
 \label{fig:6}
\end{figure}

\begin{table}[!t]
\begin{center}
\resizebox{\textwidth}{!}{\begin{tabular}{|l|c|c|c|c|c|c|c|c|c|c|c|c|c|c|c|c|c|c|}
\hline
\multicolumn{1}{|c|}{} & \multicolumn{9}{c|}{${X}_{M}$} & \multicolumn{9}{c|}{${X}_{S}$}\\
\cline{2-19}
\multicolumn{1}{|c|}{} & \multicolumn{3}{c|}{$0.75 < d_{l,r} < 1$} & \multicolumn{3}{c|}{$0.85 < d_{l,r} < 1$} & \multicolumn{3}{c|}{$0.95 < d_{l,r} < 1$} & \multicolumn{3}{c|}{$0.75 < d_s < 1$} & \multicolumn{3}{c|}{$0.85 < d_s < 1$} & \multicolumn{3}{c|}{$0.95 < d_s < 1$}\\
\cline{2-19}
\multicolumn{1}{|c|}{} & ${Y}_{0}$ & ${Y}_{M.1}$ & ${Y}_{M.2}$ & ${Y}_{0}$ & ${Y}_{M.1}$ & ${Y}_{M.2}$ & ${Y}_{0}$ & ${Y}_{M.1}$ & ${Y}_{M.2}$ & ${Y}_{0}$ & ${Y}_{S.1}$ & ${Y}_{S.2}$ & ${Y}_{0}$ & ${Y}_{S.1}$ & ${Y}_{S.2}$ & ${Y}_{0}$ & ${Y}_{S.1}$ & ${Y}_{S.2}$ \\
\hline
$\Psi_T$   & 0.67 & 0.85 & 0.85& 0.78 & 0.86 & 0.86 & 0.64 & 0.79 & 0.82 & 0.85 & 0.85 & 0.97 & 0.80 & 0.86 & 0.93 & 0.79 & 0.84 & 1\\
$\Psi_F$ & 0 & 0 & 0 & 0 & 0 & 0 & 0 & 0 & 0 & 0 & 0 & 0 & 0 & 0 & 0 & 0 & 0 & 0  \\
$\delta$ & 0.45 & 0.57 & 0.57 & 0.53 & 0.60 & 0.60 & 0.4 & 0.54 & 0.57 & 0.54 & 0.59 & 0.69 & 0.53 & 0.55 & 0.64 & 0.61 & 0.63 & 0.74 \\
\hline
\end{tabular}}
\end{center}
\caption{The median values of each boxplot displayed in \autoref{fig:6}}
\label{table:10R_S_Med}
\end{table}

\autoref{fig:6} displays the $\Psi_T$, $\Psi_F$ and $\delta$ for the zeroth (denoted $Y_0$), first, and second order AAPD models, provided with artificial antigen repertoires $Y_{M.1,S.1}$ and $Y_{M.2,S.2}$, respectively, operating on $X_M$ and $X_S$. For ease of reasing, the median values from \autoref{fig:6} are given in \autoref{table:10R_S_Med}.

\autoref{fig:6} reveals that, for both motor and sensor degradation, there is little impact from the coverage of $d_{l,r,S}$ among $R_{2-10}$ on $\Psi_T$, $\Psi_F$, or $\delta$. The only exception to this is in \autoref{fig:6}Fi-ii, where there is a slight improvement in $\delta$ and $\Psi_{T,F}$ where $R_{2-10}$ are initialised with $0.95 < d_S \leq 1$. One explanation for this that the change in motor output and sensing range is lowest in the range $0.75 < d_{l,r,S} < 1$. This also provides explanation as to why the swarm foraging performance declines once $d_0 < 0.7$ in \autoref{fig:pfddr_opti}. Interestingly, this is the range where the $\Delta P$ is the most responsive to changes in $d_{l,r}$. There is thus an opportunity for $\Delta P$ to provide a more effective means of early fault detection, but the present insensitivity of the AAPD model to $\Delta P$ alone suggests that it may need greater weighting in the matching specificity calculations (\autoref{equ:ok_spec}) to achieve the desired effect.

\autoref{fig:6}A shows that the first order AAPD model operating on $X_{M}$ gives a notable improvement in $\Psi_T$ and $\delta$ from the zeroth order model in all cases, while the additional improvement offered by the second order AAPD model is much-reduced (sometimes performing identically). Contrastingly, the first order AAPD model operating on $X_{S}$ (\autoref{fig:6}B) offers a smaller improvement over the zeroth order model, but a more pronounced improvement is offered by the second order AAPD model, in terms of $\Psi_T$ and $\delta$, at the expense of higher $\Psi_F$ in a minority of robots. This behaviour is explained principally by the difference in paratope dimension of artificial antibodies in repertoire $X_{M}$ and $X_{S}$ and the difference in values of $s$ when matching artificial antibody and antigen paratopes. $X_{M}$ consists of artificial antibodies with 3 dimensional paratopes comprising $v$, $\omega$, and $\Delta P$. While these metrics are each related to one another, there are many combinations of values they can take depending on the values of $d_{l,r}$ and an element of randomness in the sign of $\omega$ (normalised to $< 0.5$ or $> 0.5$) according to the direction of turn. This means there is overall more variation in the paratopes of artificial antibodies contained in $X_{M}$ than in $X_S$, which typically results in reduced self-stimulation of any given paratope in any given behavioural window, $W$, on average. The additional stimulation provided by ${Y}_{M.1}$, even with a relatively low value of $s = 1.5$ that limits additional stimulation only to those paratopes with very close matches, means that artificial antibody populations that may have been insufficiently exhibited for the zeroth order AAPD model to stimulate are now stimulated by the first order model. However, because of the scope for variation within ${X}_{M}$, the strict matching specificity requirements imposed by $s = 1.5$, and the fact that the AAPD computation only updates updates after 50 seconds of simulated time have passed, it is largely the same artificial antibody populations that are stimulated above threshold $F$ at the same computation cycle by the second order AAPD model, and so there is little added benefit.
On the other hand, ${X}_{S}$ contains artificial antibodies with 1 dimensional paratopes containing only $\gamma$. Although $\gamma$ can vary quickly as robots pass in and out of sensing range of one another, it does not provide ${X}_{S}$ with the stability of paratopes consisting of multiple semi-independent dimensions. Furthermore, increased values of $s$ were needed to produce satisfactory performance from the AAPD model when matching the paratopes of artificial antibody populations with $W$ and the paratopes in $Y_{S}$, with $s = 5$ and $s = 3.3$, respectively (\autoref{table:param_values}). This means that the paratopes of artificial antibodies popualtions contained in $X_{S}$ are less constrained by the requirement for a close match, resulting in the populations being additionally stimulated up to and over threshold $F$ by the AAPD model at higher values of $d_S$ across all model orders when compared to the populations contained in $X_{M}$.

\autoref{fig:6}B reveals a potential problem for single dimension paratopes and the AAPD model. The first order AAPD model, when operating on $X_{S}$, produces a relatively muted improvement in performance. This is perhaps because, as with the second order AAPD operating on $X_{M}$, the majority of artificial antibody populations that are stimulated above threshold $F$ by the first order AAPD model operating on $X_{S}$ are the same and occur at the same computation cycle as in the zeroth order model. However, there is some improvement, meaning that some new paratopes, corresponding to higher values of $d_S$, are added to repertoire $Y_{S}$. Each paratope in $Y_{S}$ will then produce a matching specificity with the paratope of any artificial antibody population in $X_{S}$ according to the value of $s$. One can imagine this in $d_S$ space, such that a paratope where $d_S = \beta$ will produce a value $m > 0$ with a subsection of paratopes froma artificial antibodies produced by robots with $d_S$ in some range $\beta \pm q(s)$, where $q$ is a function of $s$. Since paratopes in $Y_{S}$ are 1 dimensional, each time a new paratope with a higher $d_S$ is added, the base value $\beta$ of the matching space window $\beta \pm q(s)$ also increases without the counterbalances provided by additional dimensions. This is what results in the second order AAPD model operating on ${X}_{S}$ generating a relatively large number of false positives where $R_{2-10}$ are initialised with $0.75 < d_S \leq 1$ (\autoref{fig:6}B(d-i)). One possible mitigating solution to this problem would be including additional paratope dimensions -- received signal strength could be a useful candidate in this scenario. Alternatively, offsetting the additional stimulation produced by artificial antigen paratopes corresponding to learned models of faults with additional suppression corresponding to learned models of normal behaviour.

\autoref{fig:6} and \autoref{table:10R_S_Med} demonstrate the unsupervised learning ability of the AAPD model, with first and second order models improving $\Psi_T$ from as low at 64\% to as much as 86\% in ${X}_{M}$, or from 79\% to 100\% in ${X}_{S}$. Referring to \autoref{fig:5}, cases of robots being unable to return themselves to base after detection were not observed until $d_{l,r} <= 0.3$. The fact that, in this scenario, the zeroth order AAPD model operating on ${X}_{M}$ is able to keep $\delta$ in the range $0.4 \leq \delta < 0.53$ is a promising result. It is interesting that, even for first and second order models, the median average $\delta$ for true positive detections by the AAPD model could not be brought above 0.6 without generating large numbers of false positives. This, again, may be an issue with the sigmoidal fault modelling (\autoref{fig:3}) and equal weighting of paratope dimensions -- the matching of linear velocity, $v$, and angular velocity, $\omega$, which are relatively unresponsive in the range $0.75 < d_{l,r} < 1$,  contribute twice as much as the matching of the more responsive power consumption term, $\Delta P$. Nonetheless, \autoref{fig:5} shows that achieving a median $\delta = 0.6$ should be sufficient for the swarm to retain autonomy, with only a ~10\% reduction in median performance from the optimum seen at $d_0 = 0.7$. Additionally, the AAPD model operating on $X_{S}$ performs better across all orders and scenarios, and is shown in \autoref{table:10R_S_Med} to achieve the ideal desired median $\delta \approx 0.75$, $\Psi_T = 1$, and $\Psi_F = 0$ for the second order AAPD model where $R_{2-10}$ are initialised with $0.95 < d_S \leq 1$.

\subsection*{Multiple Degrading Robots \& Varying Swarm Size}

The following set of experiments examines the $\delta$ values when the AAPD model detects faults in SRS of varying sizes and where varying numbers of robots are simultaneously subjected to stochastic degradation. The AAPD model is implemented online so that, when a robot is detected as faulty, it is reset and reinitialised with its original $d_{l,r,S}$ values. This process is instantaneous (i.e. the faulty robot does not need to return itself to the base, although experiments with this physical constraint are described later in this section) in order to focus on the AAPD models response to swarms of varying size and composition. Since the number of other robots directly affects the suppression term in \autoref{equ:ok}, it is redefined proportionally to the number of robots, $N$, such that $\hat{k_2} = k_2(\frac{9}{N-1})$ where $k_2$ retains the values given in \autoref{table:param_values}, but $\hat{k_2}$ assumes the value of $k_2$ in \autoref{equ:ok} during a given experiment.

\begin{figure*}[!t]
    \includegraphics[width=\textwidth]{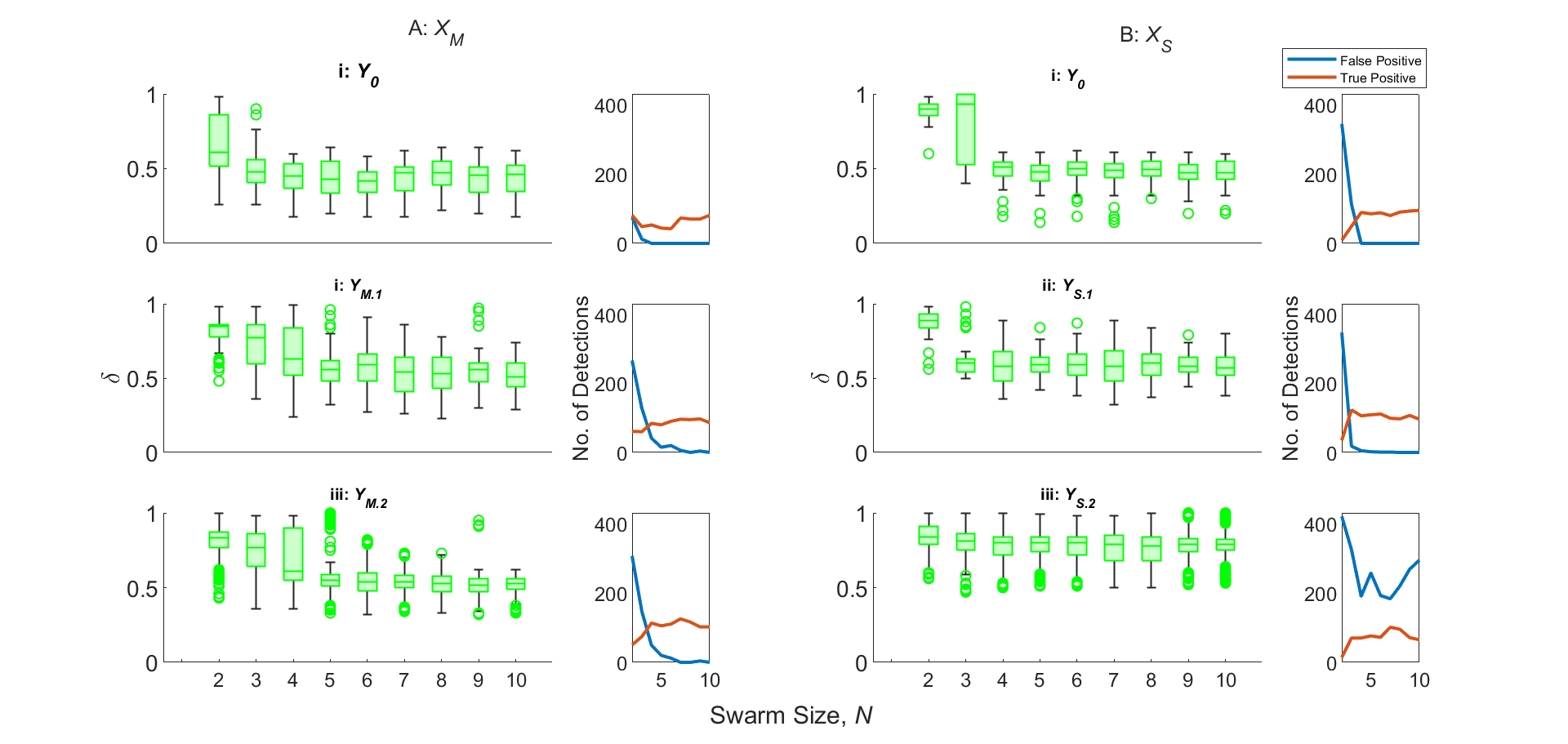}
    \caption{ $\delta$ values for robots detected as faulty by by i) zeroth (denoted ${Y}_0$), ii) first, and iii) second order AAPD models, provided with artificial antigen repertoires ${Y}_{M.1,S.1}$ and ${Y}_{M.2,S.2}$, respectively. The summed frequency of true positive (red) and false positive (blue) detections across all experimental replicates are also plotted alongside each scenario.  
    \textbf{A:} AAPD models operate on $X_{M}$ where values $d_{l,r}$ of robot $R_1$ stochastically decrease. \textbf{B:} AAPD models operate on $X_{S}$ where value $d_S$ of robot $R_1$ stochastically decreases. }  
 \label{fig:7}
\end{figure*}

\autoref{fig:7}A plots the values of $\delta$ for robots detected as faulty by zeroth, first, and second order AAPD models, in rows i, ii, and iii, respectively, operating on $X_{M}$ for SRS where $2 \leq N \leq 10$. First and second order AAPD models are provided with $Y_{M.1}$ and $Y_{M.2}$, respectively. The total frequency of true positive and false positive detections in 15 minutes of simulated time across all experimental replicates are plotted in red and blue, respectively, in the rightmost column. All robots are initialised with $P = \infty$. Robot $R_1$ is initialised with $0.9 < d_{l,r} < 1$ and a probability between 1 - 15\% of $d_{l,r}$ decrementing by 0.01 per second of simulated time, while robots $R_{2-N}$ are initialised with static $0.75 < d_{l,r} \leq 1$. \autoref{fig:7}B shows the equivalent information for the AAPD model operating on $X_{S}$ as $d_S$ decrements in $R_1$ over time with first and second order AAPD models provided with ${Y}_{S.1}$ and ${Y}_{S.2}$, respectively.

\begin{figure*}[!t]
    \includegraphics[width=\textwidth]{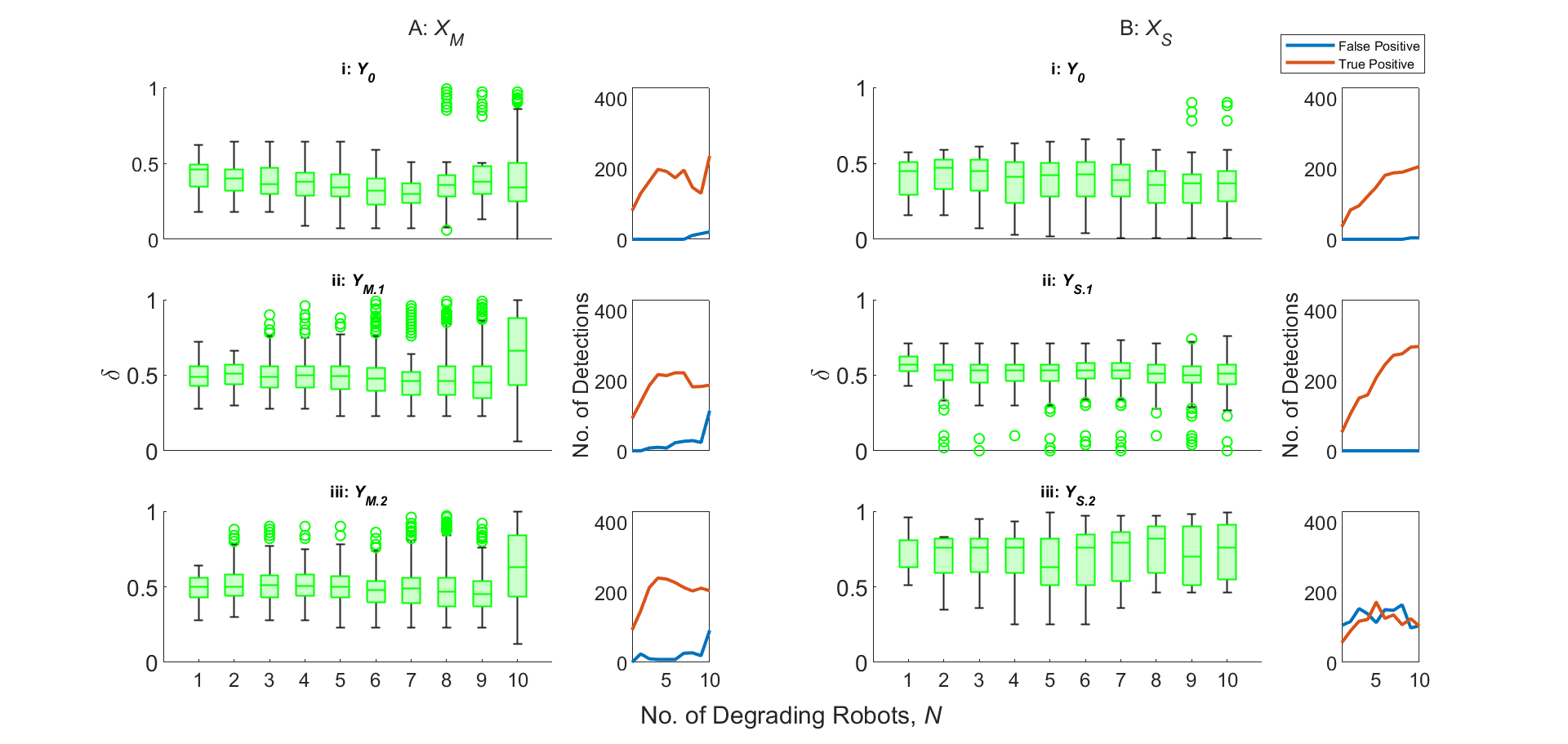}
    \caption{ $\delta$ for robots detected as faulty by i) zeroth (denoted ${Y}_0$), ii) first, and iii) second order AAPD models, provided with artificial antigen repertoires ${Y}_{M.1,S.1}$ and ${Y}_{M.2,S.2}$, respectively. The summed frequency of true positive (red) and false positive (blue) detections across all experimental replicates are also plotted alongside each scenario.  
    \textbf{A:} AAPD models operate on ${X}_{M}$ for SRS where $N = 10$ and the values $d_{l,r}$ of between 1 - 10 robots stochastically decrease \textbf{B:} AAPD models operate on ${X}_{S}$ for SRS where $N = 10$ and the values $d_S$ of between 1 - 10 robots stochastically decrease.}  
 \label{fig:8}
\end{figure*}

\autoref{fig:7} shows that the AAPD is generally insensitive to changes in $N$ above a certain point. The zeroth order AAPD model achieves a relatively consistent rate of true positive detections with false positive instances mostly eliminated for swarm sizes $N \geq 5$, beyond which adding more robots produces no consistent change to $\delta$ for both $X_{M}$ and $X_{S}$. This result demonstrates the scalable potential of the AAPD model insofar that a larger SRS would not necessarily require each robot to perform larger computations simply because there were more robots if comparable results can be achieved with a smaller sub-population. \autoref{fig:7}A shows that performance plateaus for $N \geq 5$ robots for first and second order AAPD models operating on $X_{M}$ -- although false positives are not completely eliminated for $N > 5$, the frequency of the false positives is very small above this point (i.e. less than one per experiment on average). \autoref{fig:7}B shows that the first order AAPD model operating on $X_{S}$ achieves stability for swarm sizes $N \geq 3$ and converges on this point more quickly than the zeroth order model. However, the second order AAPD model operating on $X_{S}$ never reaches a stable point and maintains a high rate of false positive detections for all swarm sizes. This effect appears to be much more severe in \autoref{fig:7}B than in \autoref{fig:6}. The reason for this is that \autoref{fig:6} shows false positives as a proportion of total experimental time, whereas \autoref{fig:7}B shows the frequency of false positive detections in a robot. This reveals that, although this AAPD model may detect as many or more false positives than true positives, it is more likely to retain true positive detections over successive computations. 

\autoref{fig:8}A plots the values of $\delta$ for robots detected as faulty by zeroth, first, and second order AAPD models operating on $X_{M}$ for SRS where $N = 10$ but multiple robots can degrade at the same time. First and second order AAPD models are provided with $Y_{M.1}$ and $Y_{M.2}$, respectively. The total frequency of true positive and false positive detections in 15 minutes of simulated time across all experimental replicates are plotted in red and blue, respectively, in the rightmost column. All robots are initialised with $P = \infty$. Between 1 - 10 robots are initialised with $0.9 < d_{l,r} < 1$ and a probability between 1 - 15\% of $d_{l,r}$ decrementing by 0.01 per second of simulated time, while any remaining robots are initialised with static $0.75 < d_{l,r} \leq 1$. \autoref{fig:8}B shows the equivalent information for the AAPD model operating on $X_{S}$ as $d_S$ decrements over time with first and second order AAPD models provided with $Y_{S.1}$ and $Y_{S.2}$, respectively. .

\autoref{fig:8} shows that the AAPD model exhibits a high degree of stability for its zeroth order implementation even where multiple robots simultaneously degrade. As one might expect, increasing the proportion of degrading robots in the swarm produces an increase in the frequency of true positive detections (although this comes with a reduction in the median $\delta$ and an increased interquartile range). The frequency of false positive detections for the zeroth order AAPD model remains consistently low. The first and second order AAPD models increase the values of $\delta$ in all scenarios. For the first order AAPD model operating on $X_{S}$, shown in \autoref{fig:8}B, this benefit comes at no expense, while the second order AAPD model, again, demonstrates its unsuitability. When operating on $X_{M}$, shown in \autoref{fig:8}A the benefits of the first and second order AAPD models are eventually offset by increased instances of false positive detection. For swarms with up to 5 simultaneously degrading robots, the frequency of false positive detections are minimal for motor degradation -- a positive result that exemplifies the promise of the AAPD model's self tolerance, learning, and memory functions. The frequency of false positive detections begins to increase, particularly for higher order AAPD models, once the majority of the swarm (6+) are simultaneously degrading, accompanied by a small decrease in true positive detections resulting from the fact that fewer robots are given the opportunity to reach $d_{l,r} < 0.75$. This is because the system no longer retains a robust majority-model of normal operation, producing a compound negative effect on the AAPD model performance. Firstly; the potential for several robots with $d_{l,r} < 0.75$ to simultaneously exhibit similar behaviour increases the suppression of corresponding paratopes and delays their detection until they occupy a sufficiently unique behavioural space. Secondly; the possibility for a majority of robots to operate with $d_{l,r} < 0.75$ and a minority with $d_{l,r} > 0.75$ means that non-faulty robots can be correctly identified as outliers, but erroneously detected as faulty. 

Overall, \autoref{fig:7} and \autoref{fig:8} show that the AAPD model is able to maintain the performance shown in \autoref{fig:6} and \autoref{table:10R_S_Med} with general stability for SRS comprising as few as 5 robots and where up to half the total population are degrading simultaneously. In general, the unsupervised selection of artificial antigen paratopes works well. However, potential problems with low paratope dimensions  are highlighted that can disrupt the unsupervised learning process if they are not addressed.

\subsection*{Diagnosis}

This study considers SRS with two classes of failure -- motor failure and sensor failure. These two classes are orthogonal in fault space for the implementation described here (i.e. a fault in a robot's motor does not directly affect the reliability of the robot's sensor and vice-versa). Therefore, when a true positive fault detection is made by the AAPD model, the robot to which the artificial antibody population $x_i > f$ belongs is identified as faulty, and can be identified as having a motor or sensor fault according to whether $x_i$ belongs to ${X}_{M}$ or ${X}_{S}$. Provided that the detection is a true positive, this information will always be correct and provides an innate diagnostic function to the AAPD model across all orders. 

More challenging is the diagnosis of faults within classes that are interactive and variable. An example of this would be diagnosing which motor has failed out of many. In the SRS studied here, this amounts to determining whether the left, right, or both motors are degraded. This is more challenging than diagnosing orthogonal faults, since each motor exerts influence on shared artificial antibody paratope dimensions. For example, $\omega \approx 0$ for motors at any stage of degradation so long as their stages of degradation are approximately equal. Similarly, two moderately healthy motors can have the same collective $\Delta P$ as a robot with one motor in perfect condition and the other severely degraded. The shape of the paratope across all dimensions is therefore critical to the sub-class diagnosis of faults. 

The proposed mechanism for sub-class diagnosis is only available to the AAPD models at the first order and above. This study assumes that, as part of the maintenance process, the true nature of any degradation can be revealed, either in supervised fashion by a trained human operator or unsupervised via diagnostic tests (e.g. \cite{kutzer2008toward, o2018fault}). Thus, paratopes in $Y$ can be associated with appropriate sub-class repair actions -- in this case, replacement of left, right, or both motors -- that provide a ground truth to the nature of the fault. Now, when a first order AAPD model detects a motor fault, the paratope in $Y$ that provides the highest matching specificity also provides a diagnosis.

To test this capability, the paratopes contained in $Y_{M.1}$ are labelled with the following categories:

\begin{itemize}
    \item Category 1: Both motors degraded ($d_l \leq 0.75$ and $d_r > 0.75$).
    \item Category 2: Left motor degraded ($d_l \leq 0.75$ and $d_r > 0.75$).
    \item Category 3: Right motor degraded ($d_l > 0.75$ and $d_r \leq 0.75$).
    \item Category 4: False positive detection ($d_l > 0.75$ and $d_r > 0.75$).
\end{itemize}

A SRS of $N = 10$ robots, $R_{1-10}$, perform the GPF algorithm in the open environments for 15 minutes of simulated time. All robots are initialised with $P = \infty$. $R_1$ is initialised with $0.9 < d_{l,r} < 1$ with independent probabilities between 1-15\% of $d_{l,r}$ decrementing by 0.01 per second, while $R_{2-10}$ are intialised with static $0.75 < d_{l,r} < 1$. The first order AAPD model is implemented online, such that robots detected as faulty are reinitialised in the base. Each time the AAPD model detects a fault, it uses $Y_{M.1}$ to perform a diagnosis.

\begin{figure}[!t]
  \centering  
    \includegraphics[width=0.7\textwidth]{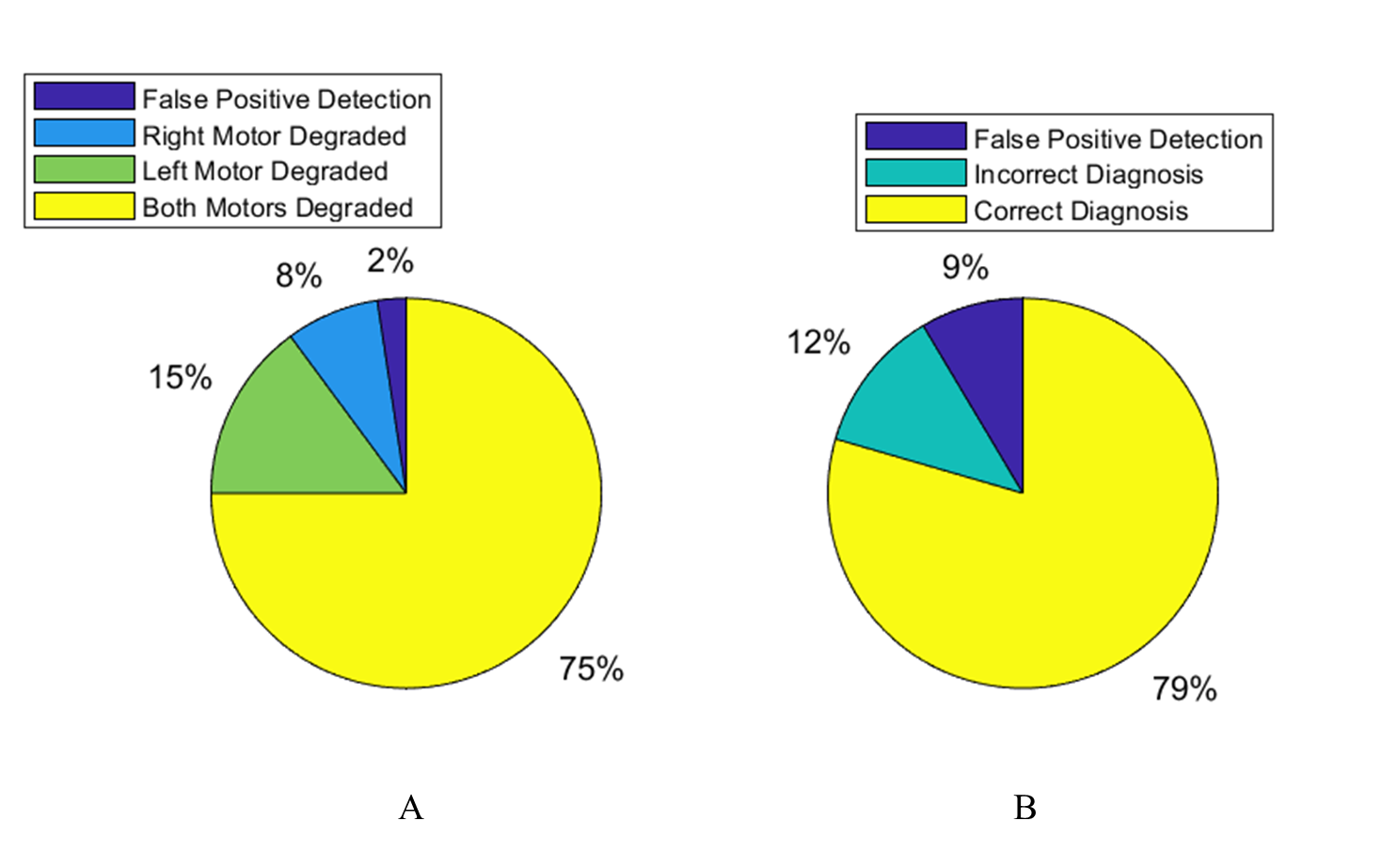}
    \caption{\textbf{A:} The true categories of faults detected by the first order AAPD model (provided with $Y_{M.1}$) operating on $X_{M}$ as a proportion of all faults detected over 10 experimental replicates. \textbf{B:} The correct and incorrect sub-class diagnoses by the first order AAPD model using $Y_{M.1}$ as a proportion of all diagnostic attempts over 10 experimental replicates.}  
 \label{fig:9}
\end{figure}

\autoref{fig:9}A displays a pie chart representing the true makeup of detection categories made over 10 experimental replicates. \autoref{fig:9}B shows the rates of correct/incorrect diagnoses made by the AAPD model over the same replicates.

\autoref{fig:9}A shows that 75\% of all detections made by the AAPD model occur when both motors are faulty. This is to be expected given that both $d_l$ and $d_r$ degrade simultaneously and \autoref{table:10R_S_Med} shows that the average $\delta$ at the moment of detection in this scenario is ~0.57, by which point it is common that both $d_{l,r} \leq 0.75$. Nonetheless, a substantial minority is accounted for by single motor failure, with only a very small proportion of false positives.  

\autoref{fig:9}B shows that, in 79\% of all cases, the paratope in $Y$ that most strongly matches with the paratope of an artificial antibody population for which $x_i > F$ yields a correct diagnosis, with only 12\% being incorrect. In fact, 79\% is an underestimate, since 9\% of all diagnoses were false-positive detections. Removing these instances from consideration, the proportion of correct diagnoses becomes ~87\%, with 13\% incorrect. The reason that false-positive detections account for 9\% of diagnoses but only 2\% of total detections is that many instances of true positive detection occur in the absence of a matching paratope in $Y$, and so they cannot be diagnosed or shown in \autoref{fig:9}B.

\subsection*{Spontaneous Faults and Environmental Variations}

Assessment of the AAPD model has so far focused on gradual degradation in robots. While this is an important mode of failure to consider, there are nonetheless scenarios in which failure can occur spontaneously or in which sudden onset environmental changes can produce similar effects.

The first set of experiments in this section considers the types of spontaneous electro-mechanical failure studied in previous swarm fault tolerance literature. The implementation of these faults is as follows:

\begin{itemize}
    \item Complete failure of both motors, $H_1$: $d_l = 0$ and $d_r = 0$
    \item Complete failure of a single motor, $H_2$: $d_l = 0$ while $0.75 < d_r \leq 1$
    \item Complete failure of sensor, $H_3$: $d_S = 0$
\end{itemize}

In the following experiments, SRS of sizes $2 \leq N \leq 10$ performing the GPF algorithm in the open environment for 15 minutes of simulated time are studied. All robots are initialised with $P = \infty$. Robot $R_1$ is initialised with one of the fault types $H_{1-3}$, while robots $R_{2-N}$ are initialised with $0.75 < d_{l,r,S} < 1$. The robot data collected is used to test the performance of the AAPD model offline.

The zeroth order AAPD model is used to produce new repertoires $Y_{M.C}$ and $Y_{S.C}$. $Y_{M.C}$ contains paratopes produced by robots detected by the zeroth order model as suffering $H_1$ or $H_2$ over 10 experimental replicates, and contains a total of 11 unique paratopes. $Y_{S.C}$ contains paratopes produced by robots detected by the zeroth order AAPD model as suffering $H_3$ over 10 experimental replicates, and contains a total of 14 unique paratopes. The vast reduction in the number of paratopes in $Y_{M.C,S.C}$ compared with $Y_{M.1,M.2,S.1,S.2}$ is indicative of the relative simplicity of complete failures $H_{1-3}$. Second order AAPD models were found to perform identically to first order models when detecting $H_{1-3}$, and are therefore not included.

\begin{figure}
  \centering  
    \includegraphics[width=\textwidth]{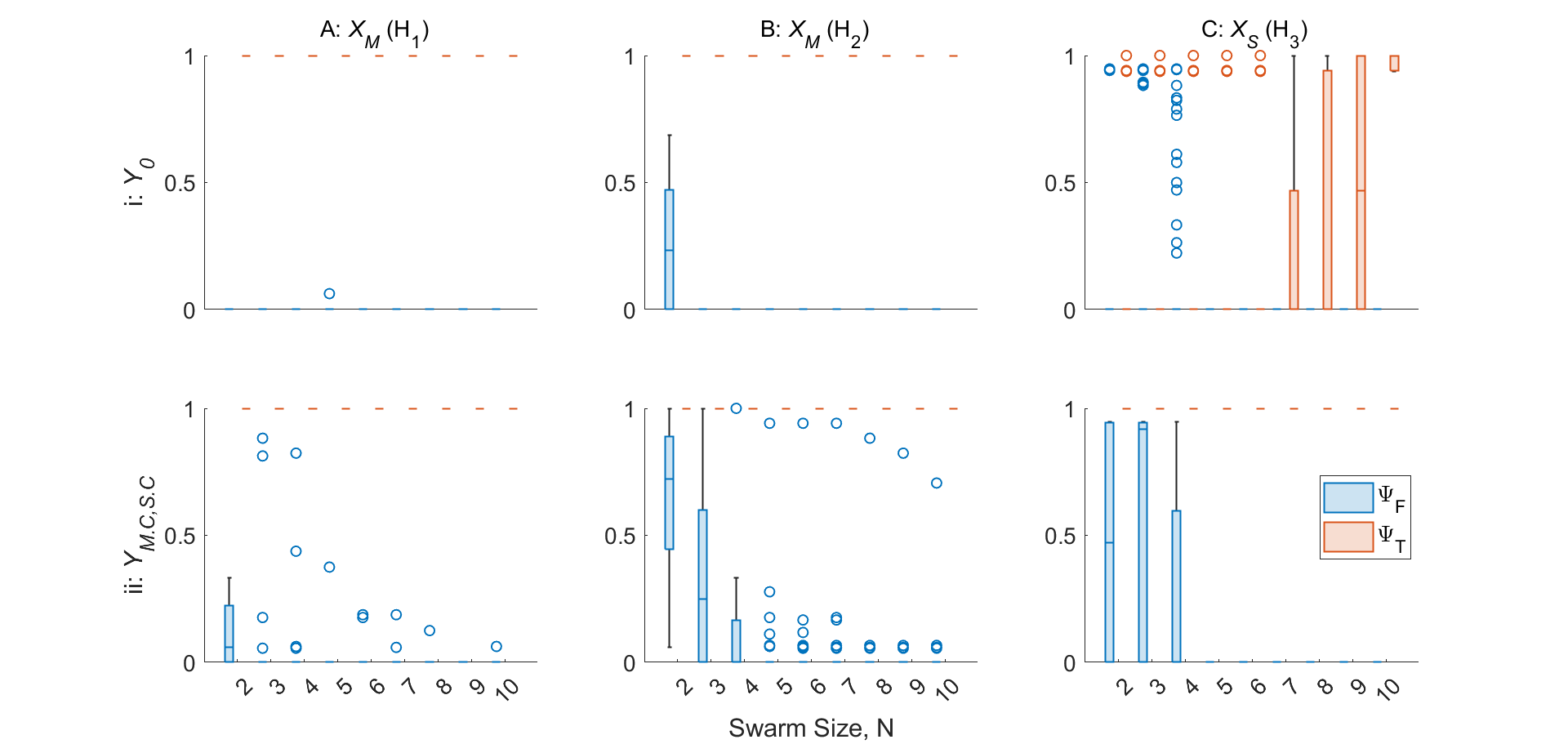}
    \caption{ AAPD model performance, $\Psi_T$ and $\Psi_F$, for varying SRS sizes, $N$, where robot $R_1$ is initialised with complete failure type ($H_{1-3}$) and $R_{2-N}$ are intialised with $0.75 < d_{l,r,S} \leq 1$. \textbf{A:}  Zeroth (i) and first order (ii) AAPD models operating on $X_{M}$ where $R_1$ suffers $H_1$. \textbf{B:} Zeroth and first order AAPD models operating on $X_{M}$ where $R_1$ suffers $H_2$. \textbf{C:} Zeroth and first order AAPD models operating on $X_{S}$ where $R_1$ suffers $H_3$.}   
    
 \label{fig:10}
\end{figure}

\autoref{fig:10} plots $\Psi_T$ and $\Psi_F$ for the zeroth and first order AAPD models operating on $X_{M}$ and $X_{S}$ with varying $N$. \autoref{fig:10}A-B show that the zeroth order AAPD model operating on $X_{M}$ is able to correctly detect robots with $H_1$ or $H_2$ for the entire experiment duration and for all values of $N$ tested, leaving no room for improvement by the first order model. The first order AAPD model remains almost completely tolerant of non-faulty robots for $N \geq 5$, while the zeroth order achieves the same for $N \geq 3$. 
\autoref{fig:10}C shows that the zeroth order AAPD model operating on ${X}_{S}$ struggles to detect $H_3$ in $R_1$ with consistency until $N = 10$. The first order model provided with ${Y}_{S.C}$ thus gives a clear improvement to the detection of $H_3$. The reason for the relatively poor performance by the zeroth AAPD order at lower $N$ is that $R_1$ spends proportionally more time outside the sensing range of the other robots it needs to obtain a reading for $\gamma < r_{max}$ and so the nature of the fault is obscured from the AAPD model for a greater proportion of experimental time.

\begin{figure*}[!t]
  \centering  
    \includegraphics[width=\textwidth]{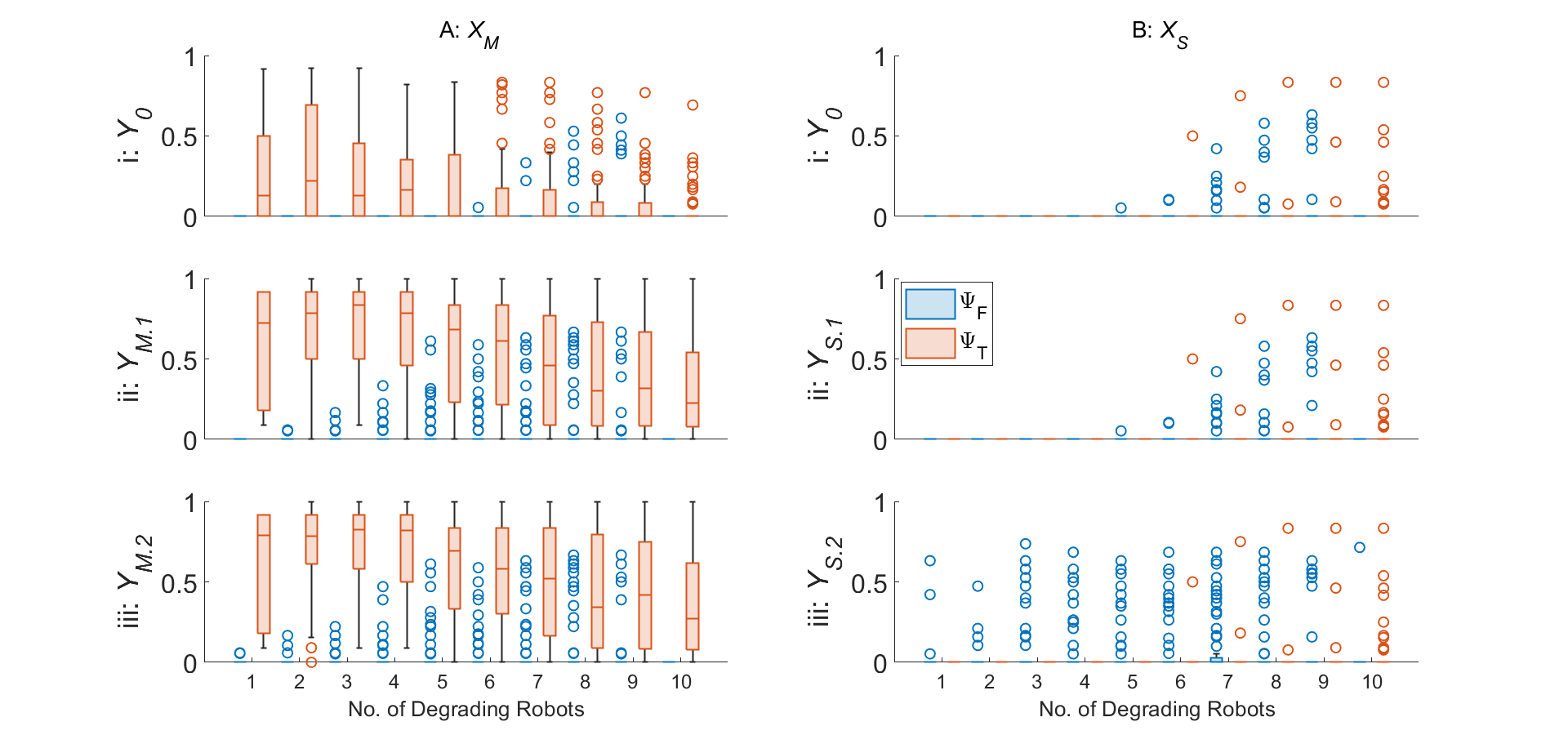}
    \caption{\textbf{A:} $\Psi_T$ (red) and $\Psi_F$ (blue) for zeroth, first, and second order (i-iii) AAPD models operating on $X_{M}$. \textbf{B:} $\Psi_T$ and $\Psi_F$ for zeroth, first, and second order AAPD models operating on $X_{S}$.  }  
 \label{fig:11}
\end{figure*}

The next set of experiments focuses on sudden but tolerable changes in robots and their environment. A SRS of $N = 10$ robots, $R_{1-10}$, perform the GPF algorithm in the open enviroment for 15 minutes of simulated time. All robots are initialised with $P = \infty$. Each robot is initialised with static $0.75 < d_{l,r,S} < 1$. After 5 minutes of simulated time, between 1 and 10 robots in the swarm have their respective $d_{l,r,S}$ values instantly degraded to $\frac{2}{3}$ of their initial value (i.e. $0.5 < d_{l,r,S} < 0.66$), where they remain until the end of the experiment (all other $d_{l,r,S}$ values are unchanged). Robot data is recorded for the duration of each experiment and used to assess the performance of the AAPD model offline. First and second order AAPD models are provided with repertoires $Y_{M.1,S.1}$ and $Y_{M.2,S.2}$ for detecting motor and sensor faults, respectively.

\autoref{fig:11}A shows that the AAPD model operating on $X_{M}$ maintains a very low $\Psi_F$ rate, with median zero across all scenarios. It can be seen that there are some outlying instances where there are false positives for relatively large proportions of experimental time, and that these are most frequent for cases where the majority (but not all) of the swarm degrades, since these are the scenarios in which the mutual suppression of remaining robots with $0.75 < d_{l,r} < 1$ will be least. For the zeroth order AAPD model, $\Psi_T$ decreases as the proportion of degraded robots increase up to and above a majority population. Where degraded robots are in the minority, the zeroth order AAPD gives a wide range of $\Psi_T$ values that appear to be substantially reduced from those presented in \autoref{table:10R_S_Med}. However, it must be remembered that \autoref{table:10R_S_Med} presents data taken from a gradually degrading robot which, once detected, will typically only become easier to detect as it degrades further. \autoref{fig:11}, on the other hand, shows data from robots that remain at $\frac{2}{3}$ of their intialised $d_{l,r,S}$ values which, in many cases, will be higher than the $\delta$ values in \autoref{table:10R_S_Med}, meaning that many of the degraded robots are tolerated by the AAPD model or are only momentarily detected as faulty. The first and second order AAPD models operating on $X_{M}$ give a higher median $\Psi_T$ across all scenarios and, in most cases, with a smaller inter-quartile range. For cases where a majority of robots degrade, however, the inter-quartile range of $\Psi_T$ remains relatively large with a low median, indicating that the majority of degraded robots are tolerated as a result of mutual suppression among the SRS.

\autoref{fig:11}B shows that the AAPD model operating on $X_{S}$ is overwhelmingly tolerant to sensor degradation of robots in the range $0.5 < d_S < 0.66$, with median $\Psi_T = 0$ and $\Psi_F = 0$ for all cases. Larger populations of degraded robots result in more outlying non-zero instances of $\Psi_T$ and $\Psi_F$  by the zeroth order model. This is most likely down to chance, where there will be a relatively small number of paratopes that a robot can exhibit that are not tolerated by the AAPD model, and the probability of these paratopes being present during any given AAPD model computation increases with increasing populations of robots with $0.5 < d_S < 0.66$. It also increases the opportunity for a relatively small number of paratopes exhibited by robots with $0.75 < d_S \leq 1$ to be detected as faulty since there are fewer robots with corresponding artificial antibody paratopes to suppress them. Such is the AAPD model's tolerance to robots with $0.5 < d_S < 0.66$ that ${Y}_{S.1}$ has no observable effect on $\Psi_T = 0$ or $\Psi_F = 0$. The unreliability of $Y_{S.2}$ has already been demonstrated in \autoref{fig:6}, and so the increased number of outlying $\Psi_{T,F}$ values are not taken to be indicative of an improved model performance.

Overall, the ability of the AAPD model to detect or tolerate the degraded robots shown in \autoref{fig:11} is neither inherently positive or negative, and model parameters were not selected with this scenario in mind. Nonetheless, what is demonstrated in \autoref{fig:11} is that the AAPD model, according to its parameter values, can detect reductions in performance affecting minority populations of robots while remaining tolerant of the same reductions affecting majority populations at the zeroth order. In a real world scenario this would potentially allow the AAPD model to tolerate substantial shifts in behaviour brought about by environmental changes that affect the majority of robots. There is ultimately a limit to the usefulness of tolerating system wide effects caused by environmental variations. If continued operation in a particular environment threatens robot autonomy, it should be identified and mitigated -- even if the entire system is affected. At higher orders, the AAPD model is able to detect reductions in performance, even where they affect majority populations, if they result in artificial antibody populations with paratopes that match sufficiently with an artificial antigen paratope. A well calibrated AAPD model, then, could potentially recognise unsustainable operating conditions and overcome corresponding mutual suppression even when a majority of robots are affected.

\subsection*{Foraging Performance}

Until now, all experiments have been conducted on SRS of $N = 10$ robots performing the GPF Algorithm (\autoref{alg_fo}) in the empty environment (\autoref{fig:2}B). However, for the final assessment of the AAPD model, the LPF Algorithm (Algorithm \autoref{alg_fo_2}), the constrained environment (\autoref{fig:setup}B), and increased swarm sizes of $N = 20$ are introduced. 

In the following experiments A SRS of $N = 10$ or $N = 20$ robots perform the GPF and LPF Algorithms for 15 minutes of simulated time. Robots $R_{1-20}$ are each initialised with $P = 1$ and independent and random probabilities between 1 - 15\% of $d_{l,r,S}$ decrementing by 0.01 per second of simulated time. 

\begin{figure}[!t]
  \centering  
    \includegraphics[width=\textwidth]{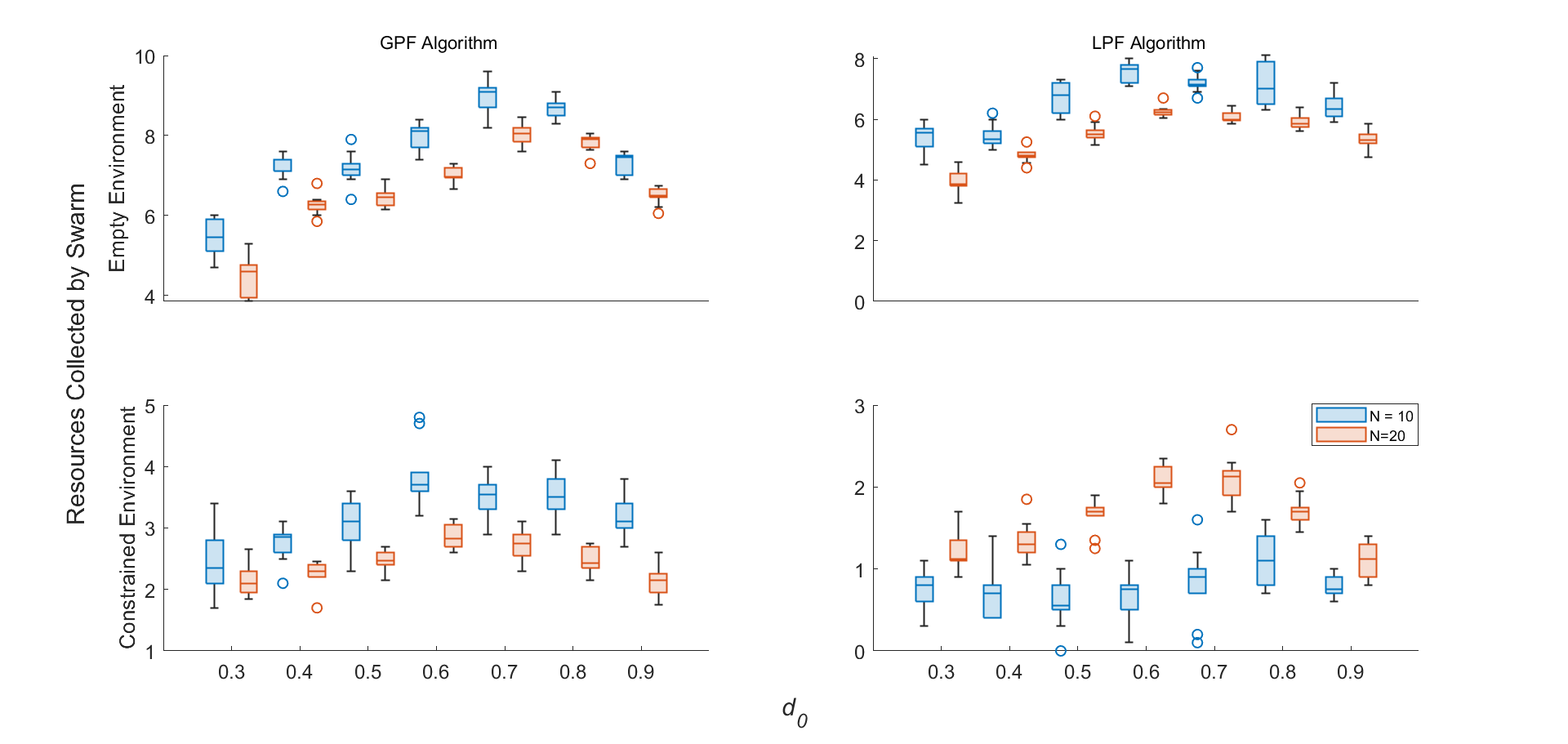}
    \caption{The resources collected (normalised) in 15 minutes by a SRS in each combination of algorithm, environment, and swarm size. Each robot degrades stochastically, with faults detected when any value $d_{l,r,S} < d_0$.}  
 \label{fig:12}
\end{figure}

Since new experimental parameters are being introduced, a baseline performance is first established whereby faults are detected in robots with any value $d_{l,r,S} < d_0$. Faulty robots return to base, where their values $d_S$ or $d_{l,r}$ are reset to their initialised values according to the type of fault detected. This is the same type of experiment as previously performed when establishing the baseline performance in \autoref{fig:3}. The median number of resources collected by each robot for each scenario combination is plotted in \autoref{fig:12}.

\autoref{fig:12} shows that the trends previously observed in \autoref{fig:3} are similarly produced for every combination of behaviour, environment, and swarm size tested. In every configuration, there is an optimum value of $d_0$, typically in the range $0.5 \leq d_0 \leq 0.8$.

The best performing AAPD models for motor and sensor faults, second and first order models with repertoires $Y_{M.2}$ and $Y_{S.1}$, respectively, are now deployed on the SRS in each scenario type. These models were shown to be able to detect motor degradation with a median average $\delta = 0.63$, albeit with a larger than desired interquartile range, and detect sensor degradation with a median average $\delta = 0.52$. When a fault is detected by the AAPD model, the robot must return itself to base in order to be redeployed with its $d_{l,r}$ or $d_S$ values reset to their initial values, according to what type of fault was detected. Note that, because of the minimum $N = 5$ robots required to run the AAPD model reliably (as shown in \autoref{fig:7}), the AAPD model will only compute on ${X}_M$ of a robot performing the LPF algorithm if it and at least 4 other robots are moving simultaneously. 

\begin{figure}[!t]
  \centering  
    \includegraphics[width=\textwidth]{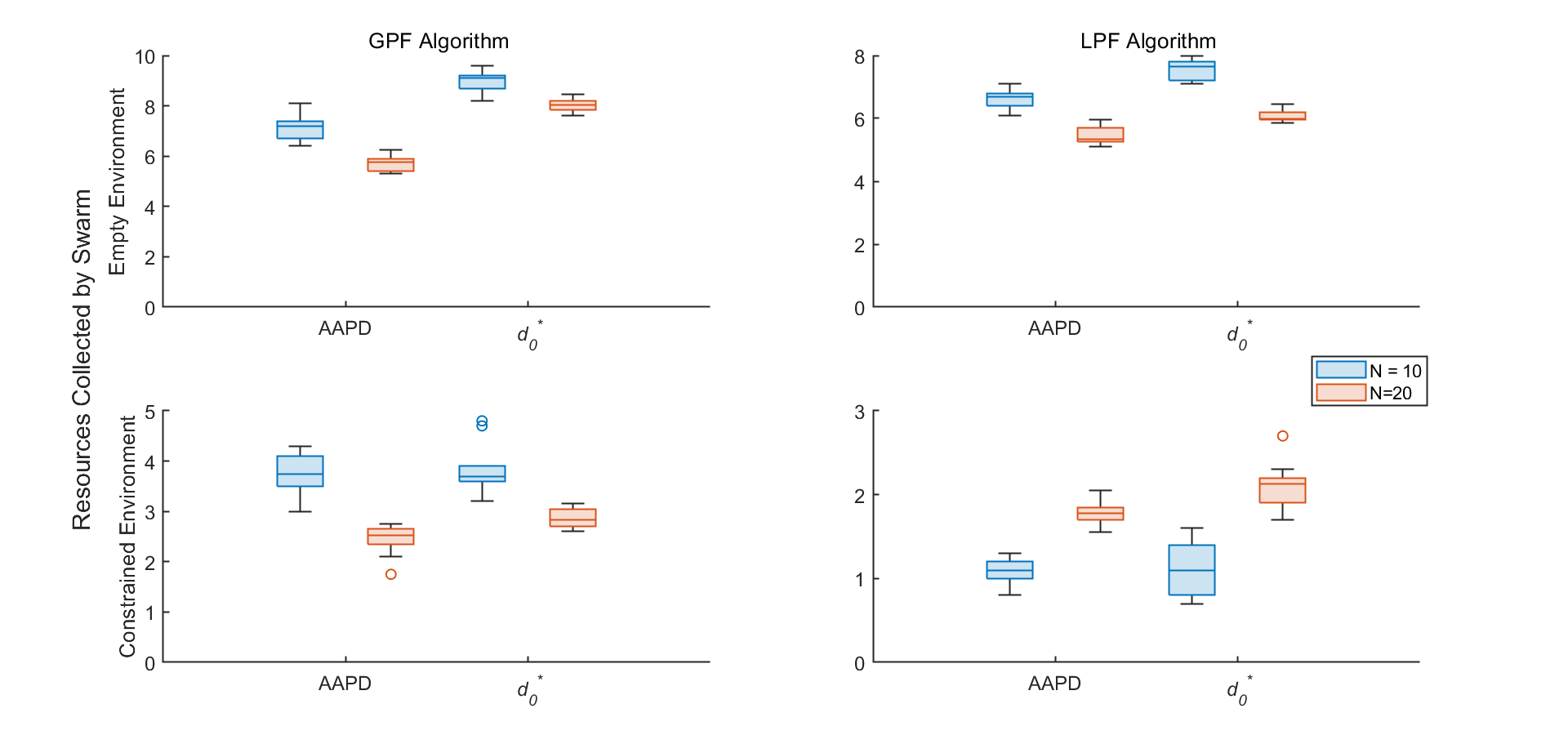}
    \caption{ The resources collected (normalised) in 15 minutes by a SRS in each combination of algorithm, environment, and swarm size. Each robot degrades stochastically. A comparison is shown for SRS performance where faults are detected by the AAPD model against faults detected when $d_{l,r,S} < d_0*$ where $d_0*$ is the best performing value of $d_0$ shown in \autoref{fig:12}.}  
 \label{fig:13}
\end{figure}

\autoref{fig:13} plots the SRS performance in each scenario when the AAPD model is implemented alongside the SRS performance with the optimal value $d_0$ taken from \autoref{fig:12}, denoted as $d_{0}^*$. For ease of reading, the median value of SRS performance when the AAPD is implemented is given as a percentage of the performance for $d_{0}^*$ in \autoref{table:AAPD}.

\begin{table}[!t]
\begin{center}
{\begin{tabular}{|p{1.5cm}|p{1.5cm}|p{1.5cm}|p{1.5cm}|p{1.5cm}|}
\hline
\multicolumn{1}{|c|}{} & \multicolumn{2}{c|}{Empty Environment} & \multicolumn{2}{c|}{Constrained Environment}\\
\cline{2-5}
 & \multicolumn{1}{|c|}{GPF} & \multicolumn{1}{|c|}{LPF} & \multicolumn{1}{|c|}{GPF} & \multicolumn{1}{|c|}{LPF}\\
\hline

N = 10 & 80\% & 89\% & 97\% & 95\%  \\
\hline
N = 20 & 70\% & 90\% & 85\% & 85\% \\
\hline

\hline
\end{tabular}}
\end{center}
\caption{The proportional difference (as a percentage) in median performance achieved by $T_1$ when compared to $T_2$ and $T_2^*$, taken from \autoref{fig:13}.}
\label{table:AAPD}
\end{table}

\begin{figure}[!t]
  \centering  
    \includegraphics[width=\textwidth]{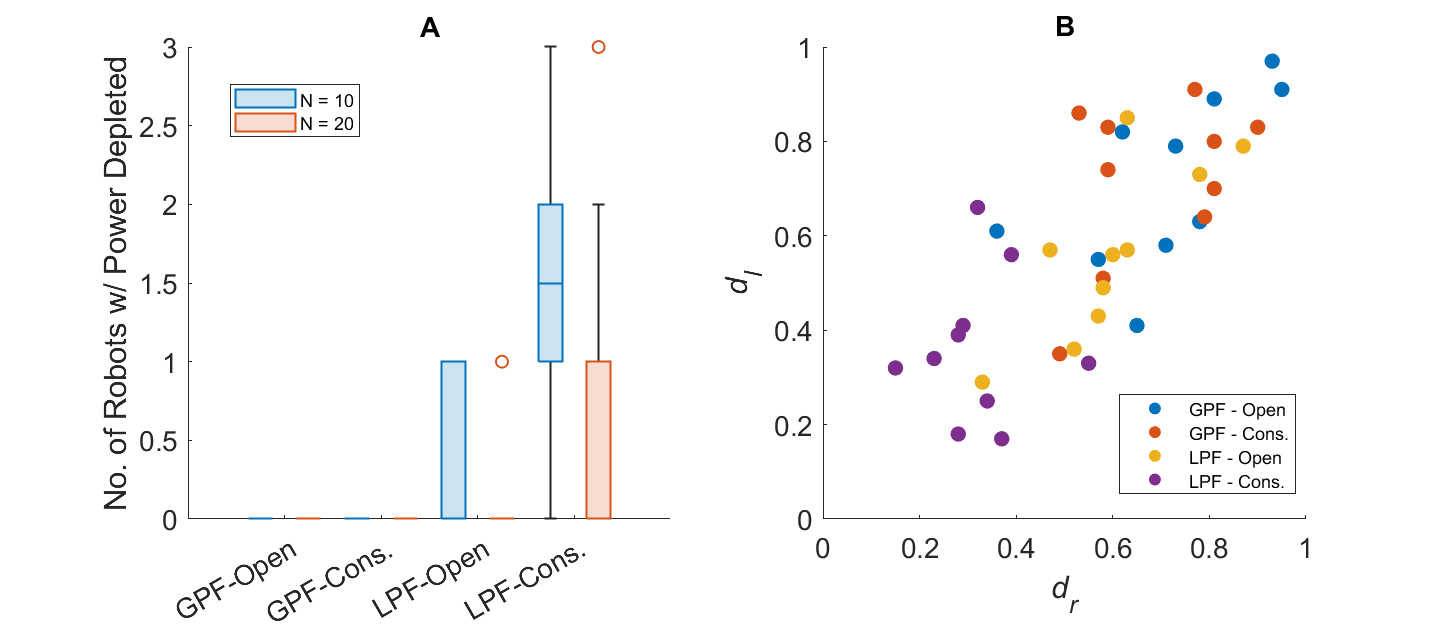}
    \caption{\textbf{A}: The number of robots that deplete their power outside of the robot base for each combination of algorithm, environment, and swarm size. Robots degrade stochastically, with faults detected by the AAPD model. \textbf{B}: The $d_{l,r}$ values of a SRS of $N = 10$ robots taken at the end of a single experimental replicate in each combination of algorithm and environment.}  
 \label{fig:14}
\end{figure}

\autoref{fig:13} and \autoref{table:AAPD} show that the AAPD model is able to give a competitive performance with the SRS performance for $d_0^*$ in many cases and, in the worst case, still enables the SRS to perform at 70\% of its optimum level. The worst performing scenario, the SRS performing the GPF algorithm in the empty environment, unsurprisingly corresponds to the highest optimum value of $d_0$ seen in \autoref{fig:12} (i.e. the furthest away from the actual median average $\delta = 0.63$ and $\delta = 0.52$ achieved by the AAPD models). Comparatively, the AAPD model achieves 97\% of the SRS performance for $d_0^*$ when performing the GPF algorithm in the constrained environment. This is also corresponds to the data plotted in \autoref{fig:12}, which places $d_0^* \approx 0.6$. Lower performances by the AAPD, in the 80\% - 90\% range, are observed for other scenarios that also have $d_0^* \approx 0.6$ in \autoref{fig:14}. This indicates the negative impact of the large interquartile range on either side of the median $\delta = 0.63$ in \autoref{fig:10}A-iii. 

In addition to the number of resources collected, $\delta$, and $\Psi_{T,F}$, an important and interrelated key performance indicator for the AAPD model is its ability to prevent robots becoming unrecoverable -- i.e. degrading to a point where they are unable to return themselves to the robot base. \autoref{fig:14}A shows the median number of robots that have completely depleted their power after 15 minutes of simulated time in each experimental scenario where faults are detected with the AAPD model. \autoref{fig:14}B shows the final $d_{l,r}$ values of each robot (as these are the key determinants of a robot's ability to complete a return to base journey) taken from an experimental replicate of each scenario where $N = 10$ and faults are detected with the AAPD model.

\autoref{fig:14}A shows that the AAPD model is able to prevent the loss of any robot in the field for 15 minutes of simulated time in all cases where SRS of $N = 10$ or $N = 20$ perform the GPF algorithm in empty or constrained environments. The median number of robots that deplete their power by the end of an experiment increases for SRS performing the LPF algorithm, but remains very low as a proportion of the swarm size $N$. The key reason for the increased number of robots depleting their power outside of the robot base when the SRS performs the LPF algorithm is that robots must necessarily be stationary for portions of experimental time. This has a dual effect of obscuring degradation on robot motors for robots that have not yet been detected by the AAPD model, and in some cases preventing the return of robots that are detected by the AAPD model if they do not have an uninterrupted communication chain to the base. In the case of SRS of $N = 10$ in the constrained environment, the ability of the SRS to provide the network coverage needed to perform the LPF algorithm is stretched to its limit, which is why SRS of $N = 20$ are less likely to experience robots that deplete their power outside of the base. The reason why the number of robots that deplete their power remains so low as a proportion of SRS size $N$ is that stationary robots only consume, at most, 20\% of the power of a robot moving in a straight line. Recalling that robots are initialised with enough power to move with uninterrupted $v_{l,r} = v_{max}$ for 5 minutes of simulated time, a robot that spent an entire experiment duration consuming power at 20\% of this rate would not have fully depleted its supply at the end of 15 minutes of simulated time. This plays a part in obscuring the susceptibility of the SRS performing the LPF algorithm to the loss of robots in the field. To better examine this effect, \autoref{fig:14}B shows the final $d_{l,r}$ values of each robot ($N = 10$) in a randomly selected experimental replicate for each scenario. 

\autoref{fig:14}B shows that, despite the low number of robots that deplete their power outside of the robot base while performing the LPF algorithm, robots commonly end experiments with $d_{l,r}$ values far below the desired level for detection -- particularly in the constrained environment. This illustrates that the long-term autonomy of the SRS is unlikely to be preserved by the AAPD model in these scenarios. However, it should be acknowledged that the implementation of degradation during the LPF algorithm, which applies a constant likelihood of degradation to a given motor irrespective of whether or not it is in use, is particularly harsh. Gradual degradation, as highlighted in Carlson's study \cite{CarlsonMTBF}, is a product of continued use and so it is obviously unlikely to occur at the same rate when robot actuators are not being used. If the rate of hardware degradation dropped to zero when an actuator was not being used, for example, one could expect to see reduced (if any) disparity in the number of robots that fully deplete their power outside of the base for SRS performing LPF or GPF algorithms, since the rate of decrement of $d_{l,r}$ would then be proportional to the distance travelled by each wheel for both algorithms. Examining the $d_{l,r}$ values at the end of an experiment for robots performing the GPF algorithm in \autoref{fig:14}, one can see that the AAPD model is able to maintain robots at a higher operational level -- with the most degraded robot in the open environment scenario finishing with $d_l = 0.41, d_r = 0.65$, and in the constrained environment with $d_l = 0.35,d_r = 0.49$ in the experimental replicate displayed. These $d_l$ values are still lower than desired, and reflect the room for improvement in the $\delta$ values of the AAPD model in its current implementation. However, referring to \autoref{fig:2}, a robot with $d_{l,r}$ at these values is likely to be able to successfully complete its return to base journey. This provides some explanation as to why there are no instances of robots depleting their power while performing the GPF algorithm, despite the higher rate of power consumption incurred by constantly mobile robots. It should be noted, however, that the $d_{l,r}$ values of robots are seen to drop below the values shown in \autoref{fig:14} on occasion, even for the GPF algorithm in the empty environment. This can be seen in the $\delta$ values plotted in \autoref{fig:10}A-iii.

Overall, the experiments conducted in this section show that the AAPD model can be applied to SRS of varying sizes operating in a variety of behavioural and environmental scenarios. Across these scenarios, the AAPD model is able to sustain the SRS performance, in terms of the median number of resources collected over the course of an experiment, at between 70\% - 97\% of the theoretic optimum level. Where the SRS performs the GPF algorithm, the AAPD model is able to detect faults such that no robot ever depletes its power outside of the robot base during an experiment, with the values of $d_{l,r}$ recorded at the end of an experimental replicate suggesting that this ability to preserve the autonomy of robots could be sustained for longer periods of time -- particularly in the open environment. It should be remembered that, although there is room to improve the results shown in \autoref{fig:13} and \autoref{fig:14}, the test-case employed here, in which all robots of a SRS are simultaneously degrading at accelerated rates, is a harsh and exaggerated scenario that is designed to test the limits of the AAPD model. That the AAPD model is nevertheless able to prevent robot failure in the field in some of these cases and allow the SRS to perform close to or at the same levels that can be seen with an ideal fault detection mechanism is an important achievement for fault tolerant swarm research.

\subsection*{Comparisons with Other SRS Fault Detection and Diagnosis Models}

The focus on faults occurring by gradual degradation in this work, and the novelty of this type of fault modelling in SRS research, means that the performances of the AAPD model in the scenarios examined so far cannot be directly compared with the performances of existing approaches to SRS fault detection and diagnosis in a 1:1 test. However, there are some fault detection scenarios in which informal comparisons can be made between the performance of the AAPD model and existing models.

The closest comparison that can be made in this work is the set of experiments studying complete motor or sensor failure ($H_{1-3}$) in a single robot while the remaining robots in the SRS (where $N = 10$) performs the GPF algorithm in the open environment, plotted in \autoref{fig:10}. This scenario is similar to the scenarios examined in \cite{taraporePLOS, carminati2024distributed}, in which a single robot from a SRS ($N = 20$) is injected with complete failure of one or both motors, denoted in \cite{taraporePLOS, carminati2024distributed} as `lact/ract' or `bact', respectively, or complete sensor failure, denoted as `pmin'. Carminati et al. \cite{carminati2024distributed} compare their proposed ML-B and ML-N fault detection models with Tarapore et al.'s \cite{taraporePLOS} CRM-B model while a SRS performs homing, aggregation, flocking, and dispersion behaviours. Performance is given in terms of an $F_\beta$-score and plotted in Figure 1 of \cite{carminati2024distributed} (details of the $F_\beta$ calculation are provided in section 3.3.3 of \cite{carminati2024distributed}). Of the SRS behaviours studied in \cite{taraporePLOS, carminati2024distributed}, dispersion, plotted in Figure 1B of \cite{carminati2024distributed}, is the closest to the GPF algorithm used in this work, in which robots perform an identical dispersion behaviour until a resource nest is located.

Comparing the AAPD model performance plotted in \autoref{fig:10} with the performance of ML-B, ML-N, and CRM-B models plotted in Figure 1B of \cite{carminati2024distributed}, the zeroth order AAPD model is able to detect complete failure of individual or both motors and complete sensor failure in a single robot from a swarm of $N = 10$ robots with $\Psi_T = 1$ and $\Psi_F = 0$. Converting $\Psi_{T,F}$ to the $F_\beta$ score format used in \cite{carminati2024distributed} gives the AAPD model $F_\beta = 1$ for detecting motor and sensor failures in the cases tested, leaving no room for improvement. Both the ML-B and ML-N models also produce a median $F_\beta = 1$ score when detecting failure in one or both motors, while the CRM-B model produces a score in the region $0.8 \leq F_\beta \leq 0.85$. In the case of complete sensor failure, ML-B, ML-N, and CRM-B models produce median scores in the range $0.4 \leq F_\beta \leq 0.7$. It should also be remembered that, where the results of the ML-B, ML-N, and CRM-B models, used for comparison here, are taken from experiments where the non-faulty robots in an SRS of $N = 20$ retain uniform ideal functionality, the zeroth order AAPD model results are taken from experiments in which the non-faulty robots of an SRS of $N = 10$ are initialised with $0.75 \leq d_{l,r,S} \leq 1$. These results thus provide strong evidence that the AAPD model gives a competitive or improved performance when compared against ML-B, ML-N, and CRM-B models in the cases tested.


\section{Conclusion}
This paper presents the bio-inspired AAPD model, a novel model for autonomously detecting and diagnosing faults. The model is inspired by Farmer et al.'s model of antibody population dynamics in the natural immune system \cite{farmer1986immune}. The AAPD model is distributed, demonstrates promising scalable properties, and is implemented and in a mostly unsupervised manner on a SRS.

The AAPD model can be implemented as a zeroth order model, a purely online data driven model that exploits SRS multiplicity to construct an implicit model of normal behaviour, but can also be provided with repertoires consisting of paratopes labelled as faulty, which typically results in improved performance -- mimicking the learning and memory functions of the natural immune system -- and enables a sub-class fault diagnosis function.   

The AAPD is tested on its ability to detect gradual or sudden degradation of motor or sensor hardware, with parameters selected to get as close as possible to median $\delta = 0.75$ and $\Psi_T = 1$. In cases of gradual motor degradation, the zeroth order AAPD model can detect faults with up to median $\delta = 0.53$ and $\Psi_T = 0.78$. The first order AAPD model improves this to as much as $\delta = 0.6$ and $\Psi_T = 0.86$. In cases of gradual sensor degradation, the zeroth order AAPD model can detect faults with up to median $\delta = 0.61$ and $\Psi_T = 0.85$, improving to as much as $\delta = 0.74$ and $\Psi_T = 1$ for the second order model. The AAPD model maintains a very low rate of false positive detections for all other model orders and scenarios. Although it can be seen in \autoref{fig:10} that the second order model operating on $X_S$ makes a large number of false positive detections, \autoref{fig:8} reveals that this does not correspond to a high $\Psi_F$ in the majority of cases -- indicating that false positive detections are not sustained. The AAPD model is able to maintain a relatively stable performance, in terms of $\psi_{T,F}$ and $\delta$, when implemented on SRS with 5-10 robots where as much as the whole swarm is simultaneously degrading. However, some AAPD model orders can be implemented on SRS comprised of fewer robots, and it can be seen that performance generally improves for scenarios where greater majorities of robots are operating in the `normal range' (i.e. $d_{l,r,S} \geq 0.75$). 

The AAPD model is able to isolate any detected fault to its corresponding robot within the SRS. If the paratopes produced by robots are separated according to the hardware they correspond to (e.g., motor or sensor in this case), any detected fault can be diagnosed to the type of hardware affected. Provided that the detected fault is a true positive, diagnosis of the robot and type will always be correct. Additionally, the first order AAPD model is demonstrated to be capable of labelling paratopes to execute finer stages of diagnosis. Where the first order AAPD model detects a motor fault, according to the characteristics of the detected paratope, it is able to correctly diagnose whether the fault is in the left, right, or both motors in 87\% of the cases tested.  

The AAPD model demonstrates a strong ability to detect sudden complete failure in motors or sensor hardware, achieving consistent median average $\Psi_T = 1$ and $\Psi_F = 0$ for SRS with as few as 5 robots in all cases tested. The AAPD model also demonstrates the ability to detect or tolerate smaller sudden reductions to motor or sensor performance to varying degrees, depending on model order, the number of robots affected, and whether motors or sensor are affected. 

When deployed on a SRS in a variety of foraging task scenarios, the AAPD model is shown to be able keep SRS performance within 70\% - 97\% of the theoretic optimum. In these scenarios, there are few, if any, instances of robots failing outside of the base area.

Although direct comparisons are hard to draw between the AAPD model and existing SRS fault detection approaches, the AAPD is shown to give a competitive or improved performance where informal comparisons can be made to existing fault detection models tested in similar scenarios.

Overall, the AAPD model presented here demonstrates a robust ability to detect a variety of potential faults and hazards while remaining tolerant to robots operating in a defined normal range and largely preventing the loss of robots in the SRS. The combination of online data-driven and model-based fault detection, as well as the option for supervised and unsupervised selection and labelling of paratopes, means that there are many possible ways of configuring and implementing the model according to the needs of a given scenario. The AAPD model integrates fault detection and diagnosis in SRS for the first time and is the first model to be applied to gradual degradation of SRS hardware. Although implemented on SRS in this work, the AAPD model has potential application on many other autonomous systems where there is multiplicity -- traditional MRS, single robots with many actuators (e.g. quad, hex, octopod robots), and potentially non-robotic systems (e.g. wind turbine arrays). This work thus makes a valuable contribution to the field of swarm fault tolerance. 

\subsection*{Future Work}

Although the AAPD model performs well in many of the scenarios tested, it also leaves room for improvement in others, with opportunity for further research and exploration. Future work will examine alternative data for use in constructing paratopes and alternative ways of obtaining a matching specificity between them. Examples include trajectory matching algorithms and the use of machine vision techniques (e.g. convolutional neural networks) on paratopes encoded as Grammian Angular Fields. Also to be considered is the use of learned paratopes of normal behaviour to counterbalance the increase in false positive detections caused by repertories of faulty paratopes in some scenarios.

Future work will also move towards long-term closed-loop FDDR by considering how each process interconnects in a real-world scenario. This will include introducing diagnostic differentiation between internal hazards (e.g. hardware degradation), external hazards (e.g. adverse terrain), and byzantine behaviour (e.g. malicious agents), as well as the introduction of a severity estimation for artificial antibody populations such that robots at greatest risk of failure in the field can be prioritised in harm mitigating actions taken by the swarm.

Finally, experimentation will extend implementation of the AAPD to different types of robotic systems (e.g. Non-SRS, legged robots, etc.), and on to real robot hardware.

\section*{Acknowledgments}

This was supported by the Royal Academy of Engineering UK IC Postrdoctoral Fellowship award under Grant ICRF2223-6-121.


\bibliography{master1.bib}
\bibliographystyle{RS.bst}

\end{document}